\PassOptionsToPackage{table}{xcolor}
\PassOptionsToPackage{nosubfloats}{jmlrutils}

\documentclass[preprint]{jmlr}

\makeatletter
\renewcommand*{\@titlefoot}{}

\renewcommand{\jmlrmaketitle}{%
  \jmlrpremaketitlehook
  \def\@jmlr@authors@sep{, }%
  \par
  \begingroup
    \def\footnoteseptext{ }%
    \def\thempfn{\textsuperscript{\thefootnote}}%
    \def\thefootnote{\fnsymbol{footnote}}%
    \if@twocolumn
      \twocolumn[\@jmlrmaketitle]%
    \else
      \@jmlrmaketitle
    \fi
    \@thanks
  \endgroup
  \label{jmlrstart}%
  \ifx\@sauthor\@empty
    \settowidth{\jmlrlength}{\@evenhead}%
    \ifdim\jmlrlength>\textwidth
      \def\@shortauthor{\@firstsurname\space et al.}%
    \fi
  \fi
  \settowidth{\jmlrlength}{\@titlefoot}%
  \ifdim\jmlrlength>\textwidth
    \def\@jmlrauthors{\@firstauthor\space \emph{et al}}%
  \fi
  \jmlrmaketitlehook
  \thispagestyle{plain}
  \setcounter{footnote}{0}%
  \let\maketitle\relax \let\@maketitle\relax
  \gdef\@thanks{}\gdef\@author{}\let\thanks\@gobble
  \def\@jmlr@authors@sep{ \& }%
}
\makeatother

\makeatletter
\def\ps@jmlrtps{%
  \let\@mkboth\@gobbletwo
  \def\@oddhead{}%
  \def\@evenhead{}%
  \def\@oddfoot{}%
  \def\@evenfoot{}%
}
\makeatother

\usepackage{times}

\usepackage[utf8]{inputenc}
\usepackage[table]{xcolor}
\usepackage{xcolor}
\usepackage{hyperref}
\usepackage[nameinlink,capitalise]{cleveref}
\usepackage{bbold}
\usepackage{xspace}
\usepackage{mathtools}
\usepackage{doi}
\usepackage{tikz}
\usepackage{enumitem}

\usepackage{thm-restate}

\usepackage[T1]{fontenc}
\usepackage{indentfirst}
\usepackage{tcolorbox}
\tcbuselibrary{listingsutf8}

\usepackage{subcaption}

\usepackage{booktabs}
\usepackage{xcolor}
\newtoggle{DEBUG}
\togglefalse{DEBUG}
\newtoggle{COMMENTS}
\toggletrue{COMMENTS}
\usepackage{tcolorbox}
\tcbuselibrary{breakable, listings, skins}

\newtcolorbox{promptbox}[1]{
  breakable,
  enhanced,
  colback=blue!5!white,  
  colframe=blue!50!black, 
  title=#1,
  fonttitle=\bfseries,
  attach boxed title to top left={xshift=5mm,yshift=-2mm},
  boxed title style={
    colback=blue!50!black,
    colframe=blue!50!black,
    sharp corners
  },
  listing only,
  left=5mm,
  right=5mm,
  top=3mm,
  bottom=3mm,
}
\usetikzlibrary{arrows.meta, positioning}

\graphicspath{{common/}}
\newcommand{\E}{\mathbb{E}}

\newcommand{\Poly}{\mathrm{poly}}

\newcommand{\pr}[1]{\Pr\left(#1\right)}
\newcommand{\abs}[1]{\left|#1\right|}
\newcommand{\aldrift}{\textsc{ALDrIFT}}
\newcommand{\Lzero}{\mathcal{L}_0} 

\renewcommand{\L}{\mathcal{L}}

\newtheorem{assume}{Assumption}

\crefname{Program}{Program}{Programs}
\creflabelformat{Program}{(#2\textup{#1})#3}

\renewcommand{\epsilon}{\varepsilon}

\makeatletter
\def\@fnsymbol#1{\ensuremath{\ifcase#1\or \dagger\or \ddagger\or
   \mathsection\or \mathparagraph\or \|\or **\or \dagger\dagger
   \or \ddagger\ddagger \else\@ctrerr\fi}}
\makeatother

\definecolor{linkc}{rgb}{0.1, 0.5, 0.7}
\definecolor{citec}{rgb}{0.6, 0.3, 0.7}
\definecolor{urlc}{rgb}{0.5, 0.1, 0.2}
\hypersetup{
    colorlinks=true,
    linkcolor=linkc,
    citecolor=citec,
    urlcolor=urlc
}

\newcommand{\alift}{\textsc{ALDrIFT}\xspace}
\newcommand{\topift}{\textsc{TopIFT}\xspace}

\newcommand{\tv}{d_{\mathrm{tv}}}
\newcommand{\model}{\mathcal{L}}

\newcommand{\temp}{\tau}
\newcommand{\prevtemp}{\temp {\scalebox{0.8}[0.8] -}}
\newcommand{\maxtemp}{T}

\begin{document}

\title{Sample-Efficient Optimization over Generative Priors via Coarse Learnability}

\author{%
  \Name{Pranjal Awasthi} \Email{pranjalawasthi@google.com}\\
  \Name{Sreenivas Gollapudi} \Email{sgollapu@google.com}\\
  \Name{Ravi Kumar} \Email{ravi.k53@gmail.com}\\
  \addr Google Research, Mountain View, USA
  \AND
  \Name{Kamesh Munagala}\footnote{This work was done while the author was visiting Google Research.} \Email{kamesh@cs.duke.edu}\\
  \addr Department of Computer Science, Duke University, Durham, USA
}

\maketitle

\begin{abstract}
We study zeroth-order optimization where solutions must minimize a cost $d(s)$ while maintaining high probability under a complex generative prior $\mathcal{L}(s)$ (e.g., a parameterized model). This reduces to sampling from a target distribution proportional to $\mathcal{L}(s) e^{-T \cdot d(s)}$. Since classical model-based optimization (MBO) lacks finite-sample guarantees for expressive approximate learners, we introduce \emph{coarse learnability}, a flexible statistical assumption requiring only that a learned model covers the target's probability mass within a polynomial factor. 

Leveraging this assumption, we design an iterative MBO algorithm called \alift with a sample correction step that provably approximates the target using only a polynomial number of samples. We apply this framework to globally optimizing non-convex objectives bounded by a quadratic envelope in $\mathbb{R}^n$, where we show this assumption is naturally satisfied for a family of ``optimistic'' posterior distributions. To reach global $\varepsilon$-optimality, this implies a sample complexity of $\widetilde{O}(\log 1/\varepsilon)$, a rate characteristic of optimistic space-partitioning methods.

We further justify coarse learnability as an assumption for generative priors theoretically, proving that in simple settings, parametric maximum likelihood estimation and over-smoothed kernel density estimators naturally satisfy it. 

Finally, one motivation for our framework comes from inference-time alignment. Though our primary contribution pertains to the theoretical foundations of MBO, we provide qualitative evidence that, in simple settings, even primitive LLMs can shift their distributions toward lower-cost regions when fine-tuned with zeroth-order feedback.
\end{abstract}
\section{Introduction}

Model-based optimization (MBO) is a broadly applicable framework for zeroth-order optimization: a probabilistic model over candidate solutions is iteratively updated using function evaluations, guiding search toward high-performing solutions. Classical MBO methods such as the cross-entropy method~\citep{rubinstein1999cross} are well-studied and widely used, but their theoretical guarantees are asymptotic (as opposed to finite sample) and rely on strong structural assumptions. These assumptions break down when the generative model is an expressive approximate learner, such as a neural network, leaving the finite-sample behavior of MBO in this regime theoretically uncharacterized. As discussed more in \cref{app:extended_motivation}, a motivating instance is inference-time optimization with learned generative priors, including language models, where one seeks to enforce global objectives while retaining high prior likelihood under the generative model. While analyzing MBO over modern generative models remains  challenging, establishing finite-sample guarantees in tractable, continuous domains is a necessary first step. This requires  more flexible assumptions on the generative model, which is the focus of this paper.

\subsection{Model-Based Optimization (MBO) with a Generative Prior}
We study a zeroth-order optimization framework, where the goal is to minimize a function $d(\cdot)$ given only oracle access to its values. The function $d(\cdot)$ might represent a continuous non-convex loss surface, or a discrete penalty for violating global constraints. To guide this search, we assume access to a generative prior $\L(\cdot)$ and our objective is to sample effectively from a target distribution of the form $p_T(s) \propto \L(s) \cdot e^{-T\cdot d(s)}$, where $T$ is a `temperature' parameter. Samples from this distribution naturally balance low values of $d(s)$ with high probability under the prior $\L(s)$. This formulation is versatile: in continuous domains, the prior provides regularization to escape local minima (see \cref{app:nonconvex_mbo}); in combinatorial settings, it allows a division of labor where an external checker enforces global feasibility via $d(\cdot)$ while a generative model enforces semantic priors via $\model(\cdot)$.  

Our primary object in this paper is sampling from $p_T$; approximate optimization of $d$ follows by choosing $T$ appropriately large. For such $T$, the target distribution $p_T$ can become \emph{exponentially far} from the initial prior $\L$, making direct sampling or simple rejection-based corrections infeasible. 

\noindent {\bf Sample-Efficient Model-Based Optimization.} To address this sampling challenge, we adopt the perspective of model-based optimization (MBO), most notably the cross-entropy method~\citep{rubinstein1999cross} and model reference adaptive search~\citep{hu2007model}. (\Cref{sec:related_work,sec:mras_connection} contain a detailed comparison.) These methods, which are widely used for derivative-free non-convex optimization~\citep{shahriari_taking_2016, snoek_practical_2012}, optimize an objective by iteratively updating a probabilistic or neural search distribution that concentrates on the optimum. However, classical guarantees for these methods typically rely on asymptotic convergence arguments via structural properties of specific surrogates, such as natural exponential families~\citep{hu2007model}. These results do not extend to the finite-sample regime, either with well-structured priors or with generative models  that function as  approximate learners. Our work bridges this gap by providing polynomial sample complexity bounds for this class of algorithms under a general \emph{coarse learnability} assumption, replacing rigid structural constraints with a statistical coverage condition.

\noindent {\bf Simulated Annealing and \aldrift{}.} We operationalize this  framework via \aldrift{} (\emph{Algorithm Driven Iterated Fitting of Targets}), an iterative algorithm in which a generative model is progressively refined.  We construct an intermediate distribution $p_{\tau}(s) \propto \L(s) \cdot e^{-\tau\cdot d(s)}$ and detail how \aldrift{} treats $\tau$ as a (inverse) temperature parameter---drawing inspiration from simulated annealing~\citep{kirkpatrick1983optimization, KalaiV, jerrum2004elementary}---and gradually increases it to $T$ to assign more weight to the objective $d$ (see \Cref{sec:iterated}). Crucially, at each iteration, \aldrift{} performs a Metropolis--Hastings step to generate samples from the current target distribution $p_\tau(s)$, using the learned generative model as a proposal. In this setting, using importance~\citep{thomas2016data} or rejection sampling~\citep{huang2025inferencealignment} would require exponentially many samples. Metropolis--Hastings circumvents this by relying solely on \emph{relative} likelihood ratios, allowing the procedure to remain efficient even when the target and initial distributions differ exponentially.

The resulting samples are then used to fit the new generative model (via say parameter estimation) so that it produces samples that are more closely aligned with $p_\tau(s)$. At a high level, this approach is analogous to related methods in convex optimization \citep{KalaiV}, to approximate policy-iteration schemes from reinforcement learning~\citep{schulman2015trust,agarwal2020optimality,kakade2002approximately,peters2007reinforcement,peters2010relative}, and to sequential MCMC~\cite{SMC,PMC,neal1998annealedimportancesampling,particleMCMC}, though the details and objectives differ. Further, as discussed in \cref{sec:mras_connection}, \alift is an instance of the model reference adaptive search (MRAS) framework~\citep{hu2007model}, and our algorithm and analysis can be interpreted as a finite-sample robustification of MRAS via the Metropolis--Hastings proposal step.

\noindent {\bf Formal Guarantees via Coarse Learnability.}
Our main theoretical contribution in \Cref{sec:coarse} is a proof that \aldrift{} converges to the target distribution $p_T(s)$ with polynomial sample complexity, under a novel coverage assumption we call {\em coarse learnability}. (See \Cref{ass:learn} in \Cref{sec:coarse1} for a formal statement.) This assumption formalizes approximate learning as follows: given $m$ samples from a target distribution, we can learn a generative model whose density envelopes the target density within a factor of $\mbox{poly}(m)$, except on an exponentially small error set. 

A central conceptual contribution of our work is the formulation of coarse learnability itself. At a conceptual level, as discussed in \Cref{sec:coarse}, coarse learnability identifies a sufficient statistical condition under which any sampling-based learner can provably refine a generative model toward a target distribution.  
As we discuss in \Cref{sec:related_work}, coarse learnability is related to the notion of {\em coverage} in reinforcement learning~\citep{munos2008finite,xie2022coverage}, which similarly quantifies when a policy explores enough of the optimal state-action space to permit efficient improvement. Coarse learnability  differs by treating coverage as a dynamic statistical outcome of the model fitting instead of as a static assumption, regenerating the required coverage dynamically at each iteration.

\noindent {\bf Provable Bounds for Non-Convex Global Optimization.} \cref{sec:coarse} gives a general finite-sample guarantee for \alift under \cref{ass:learn}. In \cref{app:nonconvex_mbo}, we show the utility of this result in the classical setting of $\epsilon$-approximately globally optimizing a non-convex function in $\mathbb R^n$, where prior results only show asymptotic convergence of MBO with infinitely many samples~\cite{hu2007model}.  In contrast, we prove that on a well-studied class of quadratically bounded non-convex functions~\citep{bubeck2011x,munos2011optimistic} with known curvature bounds, \alift is sample-efficient. Unlike existing MBO methods that fit an exact exponential family model, we  fit a wider Gaussian posterior, analogous to optimistic exploration in bandit literature~\citep{bubeck2011x}. This key difference enables us to verify  \cref{ass:learn} in this setting, and show that a strengthening of \alift achieves sample complexity $\widetilde{O}(\log 1/\epsilon)$, with dimension- and geometry-dependent constants. This rate is characteristic of optimistic space partitioning methods~\citep{munos2011optimistic,bubeck2011x} (see \cref{app:nonconvex_mbo} for some caveats) and is exponentially better than that achievable by local MCMC methods~\citep{holley1987logarithmic,hajek1988cooling}. We note that for this sample-complexity guarantee, \cref{ass:learn} is no longer an assumption, as it is rigorously verified. Complementing our theory, we present an empirical demonstration of the need for the inflated proposal and sample correction steps with limited sample budgets in \cref{sec:empirical_10d}.


\noindent {\bf Theoretical Plausibility of Coarse Learnability.}
In \Cref{sec:justify}, we present theoretical evidence justifying \cref{ass:learn} for certain generative models. We demonstrate that standard MLE naturally induces the required coverage properties in several simplified settings: the {\em agnostic} setting where the model is misspecified (e.g., a unimodal learner approximating a multimodal target) and precise density approximation is impossible; for the \emph{realizable} setting for well-behaved exponential families; and  {\em non-parametric settings} such as standard Kernel Density Estimation.

\noindent {\bf Motivating Application: Inference-time Alignment.} Though our primary focus is to theoretically formalize the mechanics of MBO under generative priors, we note in \cref{app:extended_motivation} that one motivating application is the combination (via inference-time alignment and fine-tuning) of classical combinatorial algorithms with generative models~\citep{huang2025inferencealignment,chen2025sets,madaan2023selfrefineiterativerefinementselffeedback} to solve problems that require both global feasibility and adherence to local priors: route planning with qualitative scenic preferences as prior, molecular design with learned drug-likeness priors, or scheduling with semantic session coherence. In such problems, the generative model encodes local constraints that a classical algorithm cannot capture, while the algorithm enforces global structure that a generative model cannot reliably guarantee. LLMs are a natural family of priors for this purpose, though, as we illustrate in \cref{app:extended_motivation}, they can fail to enforce global combinatorial properties on their own, confirming the need for the hybrid approach.

We present qualitative results in \Cref{sec:empirical} demonstrating that even a relatively primitive LLM can iteratively acquire combinatorial constraints (initially unknown to it) from only a few samples via fine-tuning. Specifically, we show that for a simple scheduling problem and for low degree spanning trees, when guided by a heuristic version of \aldrift{} called \topift, an LLM (GPT-2~\citep{radford2018improving}) can adapt its generative distribution to favor solutions with low $d(s)$ values while still respecting its prior, hence qualitatively demonstrating the ``mass covering'' type behavior required for coarse learnability. 

While our formal guarantees do not yet apply to inference-time alignment directly (and likely require relaxed formulations), our aim is to introduce an MBO framework that makes the behavior of expressive and potentially mis-specified models more amenable to analysis.  By identifying coverage as a key driver of convergence, we provide a possible theoretical lens for reasoning about how generative models could support combinatorial optimization under appropriate coverage conditions.


\noindent {\bf Summary of Contributions.} We summarize our primary contributions as follows:
\begin{itemize}
    \item \textbf{Conceptual Framework:} We introduce \textit{coarse learnability} (\cref{ass:learn}) as a sufficient coverage-style condition to establish finite-sample guarantees for iterative Model-Based Optimization (MBO) using expressive approximate learners.
    \item \textbf{MBO Algorithm:} We propose \alift (\cref{sec:alift}), an MBO algorithm that pairs iterative learning of generative models with a principled sampling correction step (e.g., Metropolis-Hastings) to control the accumulation of approximation errors.
    \item \textbf{General Theoretical Guarantee:} We prove that under \cref{ass:learn}, \alift achieves polynomial finite-sample convergence to the target $p_T$ (\cref{thm:main_iter}).
    \item \textbf{Unconditional Instantiation:} To ground our framework, we analyze \alift in the canonical setting of $\epsilon$-globally optimizing non-convex objectives bounded by a quadratic envelope in $\mathbb{R}^d$ in \cref{app:nonconvex_mbo}. By employing an optimistic Gaussian proposal, we  show that coarse learnability holds in this setting and obtain a sample complexity of $\widetilde{\mathcal{O}}(\log 1/\epsilon)$. 
    \item \textbf{Supporting Evidence:} Finally, we provide proof-of-concept theoretical and qualitative empirical evidence (\cref{sec:justify,sec:empirical}) demonstrating respectively that certain simple generative models satisfy coarse learnability and that even primitive LLMs can exhibit the mass-covering behavior required by our theoretical framework on combinatorial tasks.
\end{itemize}

\subsection{Related Work}
\label{sec:related_work}
Our work sits at the intersection of several active research areas, including zeroth-order optimization, statistical learning theory, sequential MCMC, and non-convex optimization. We present the most pertinent related work here, and defer related work on non-convex optimization methods to \cref{app:compare_mbo}.

\paragraph{Model-Based Optimization and MRAS.} Our approach is formally an instance of the model reference adaptive search (MRAS) framework~\citep{hu2007model}, which generalizes the cross-entropy method~\citep{rubinstein1999cross}.  MRAS operates by using a parametric model to approximate a sequence of ideal target distributions that increasingly concentrate on the optimum. In each epoch, it generates samples from the current parametric model, selects the high-performing ``elite'' samples, and updates the model parameters to fit this elite set. While MRAS is proven to converge in the limit to the globally optimal solution when the parametric distributions are from an exponential family, it lacks sample complexity bounds, both for simple parametric families as well as for expressive approximate learners, which is the focus of our paper. 
We present this connection in more detail in \cref{sec:mras_connection}.

Our work also bears similarity to reinforcement learning algorithms
such as reward-weighted regression~\citep{peters2007reinforcement}
and relative entropy policy search~\citep{peters2010relative}.
\citet{huang2025sharpening} analyze language-model
sharpening through supervised fine-tuning and reinforcement learning,
assuming that the model class can represent the corresponding target
distributions. Their framework includes KL-regularized objectives whose optimal distributions have the same form as our targets. Our framework allows these targets to lie outside the learned model family. Under coarse learnability, we obtain polynomial sample-complexity bounds by using the learned models as proposals and correcting their sampling bias via Metropolis--Hastings.

\paragraph{Sequential Monte-Carlo Sampling.} The annealing structure of \alift, progressing through a sequence of tempered distributions $p_\tau(s) \propto \model(s)e^{-\tau d(s)}$ with Metropolis--Hastings corrections at each step, shares a foundational skeleton with Annealed Importance Sampling (AIS;~\citep{neal1998annealedimportancesampling}) and Sequential Monte Carlo (SMC;~\citep{SMC,particleMCMC}). However, these methods maintain a discrete empirical measure over particles, while \alift introduces a parametric projection step: it fits a generative model to the particles at each temperature, which is then used as the proposal for the next step. In effect, this replaces the variance associated with importance weights and particle degeneracy with approximation bias, as the fitted model $\model_\tau$ may not faithfully cover the target $p_\tau$. While related ideas appear in adaptive and amortized sampling methods, this projection step introduces a distinct source of error not present in classical particle-based schemes, which we address via the coarse learnability assumption.

Population Monte Carlo (PMC;~\cite{PMC}) shares a similar overall structure to \alift. It iteratively fits a parametric proposal to a particle population; however, in our setting, the variance in weights can explode. \alift circumvents this by defining a sequence of intermediate targets and utilizing Metropolis-Hastings to draw unweighted samples from them.  Consequently, our contribution is to present sufficient conditions under which such unweighted sampling has polynomial complexity.

Furthermore, while recent work on amortized MCMC via transport maps similarly focuses on ``learning to propose'' to improve acceptance rates~\citep{ParnoMCMC,Brofos2021Adaptation,hoffman2019}, we introduce a finite-sample coverage assumption (\cref{ass:learn}) rather than relying on accuracy of the learned transport map to improve sampling efficiency.

\paragraph{Learnability and Coverage.}
In contrast to classical statistical learning theory concepts such as probably approximately correct (PAC) learning \citep{Valiant},  our paper introduces the notion of coarse learnability (\cref{ass:learn}). 
This concept is loosely related to the coverage condition studied in reinforcement learning~\citep{munos2008finite,xie2022coverage,xie2024exploratorypreference} and recent work on inference-time alignment~\citep{huang2025inferencealignment}. In those settings,  bounded coverage is typically a static assumption on the data distribution or behavior policy, ensuring sufficient overlap with the optimal policy to permit valid estimation. Coarse learnability plays a similar structural role, but differs by treating coverage as a statistical outcome of the model fitting. It posits that given polynomial samples from a target, the generative model learns to cover that target (within polynomial factors), effectively regenerating the required coverage dynamically at each iteration. 

\paragraph{Bandit Convex Optimization.}
Our problem involves optimizing an objective $d(s)$ where only function evaluations are available. This is related to bandit convex optimization (BCO)~\citep{FlaxmanKM05}, which provides regret bounds for online convex minimization with only noisy function evaluations.  Conceptually, if the LLM's probability function $\mathcal{L}(s)$ were log-concave and the algorithmic cost $d(s)$ convex, the target distribution $p_\tau(s) \propto \mathcal{L}(s)e^{-\tau d(s)}$ would be log-concave, making it amenable to BCO techniques. Our framework extends this to the general scenario where the prior $\mathcal{L}(s)$ need not be log-concave, and the function $d(s)$ need not be convex. We therefore need a somewhat weaker learnability condition that is applicable to these situations.

\paragraph{Bayesian Methods.} Coarse learnability is conceptually related to posterior-contraction results in Bayesian nonparametrics~\citep{ghosal2017fundamentals,vdv2008rates}, which characterize how a Bayesian posterior concentrates around the true distribution with increasing data.  Unlike those asymptotic consistency results, our assumption is a finite-sample condition ensuring that a learned generative model maintains sufficient mass on high-value regions during iterative optimization.
Finally, our work bears superficial similarity to the use of Bayesian posteriors in guiding exploration~\citep{srinivas2010gaussian,russo2014learning}. However, in those settings the posterior merely quantifies uncertainty and could be replaced by frequentist confidence bounds such as UCB. In contrast, in our framework the generative model is an integral component of the optimization process itself, progressively refining its distribution toward the optimal solution.



\section{Optimization Framework with a Generative Model}
\label{sec:model}

We consider the problem of minimizing a global objective function $d: S \to \Re$, where $S$ is the solution space.%
\footnote{In practice, $S$ is conditioned on the prompt string $x$, which encodes the local and global constraints of the problem instance; we omit $x$ from the subsequent discussion.} As is standard in MCMC literature, we assume w.l.o.g.\ that $d(s) \in [0,D]$ for all $s \in S$. We assume only zeroth-order oracle access to $d(s)$: we can evaluate $d(s)$ for any $s \in S$, but do not have access to its gradients. In our canonical applications (e.g., \cref{sec:empirical}), $d(\cdot)$ 
can typically be computed efficiently by an algorithm. In addition, we assume access to a prior distribution $\L(s)$. This prior $\L(s)$ can be modeled by a generative distribution, such as that provided by parametrized or generative models.

Our goal is to identify solutions $s$ that attain low values of $d(s)$ while also having high probability under $\L(s)$.  This objective can be formulated as sampling from, or finding the modes of, the target distribution:
$$ p_{T}(s) \propto \L(s) \cdot e^{-T \cdot d(s)}, $$
where $T > 0$ is a `temperature' parameter that controls how sharply $p_{T}(s)$ concentrates on solutions $s$ that minimize $d(s)$ among those with significant probability mass under $\L(s)$.

This sampling problem is particularly interesting when the generative model $\L$ is very unlikely to generate solutions with small $d(\cdot)$, and when a random solution $s$ with low $d(s)$ is also very unlikely to have relatively high probability under $\L$ compared to a carefully chosen $s$. We assume that given samples from any  distribution $\mathcal{D}$, we can use these samples to learn a distribution that is close to $\mathcal{D}$, for instance, via parameter-fitting.  We will make the learning assumption more precise later, and this will be one of our key contributions. 

In the sequel, we use $s \sim \L$ to denote that the sample $s$ is generated by a model $\L$. For distributions $P, Q$, let $\tv(P, Q)$ denote the \emph{total variation distance} between them.

\section{\alift: An Algorithm with Provable Guarantees}
\label{sec:iterated}

We now develop our algorithm \alift for solving the sampling problem above, with provable guarantees. Before presenting the algorithm, we first discuss a natural heuristic, \topift, which highlights key technical challenges in obtaining formal bounds and thereby motivates both our algorithm and the assumptions underlying it.

\subsection{A Simple Heuristic: \topift}
\label{sec:topift}

To build intuition for why our sampling problem is non-trivial, we begin with a natural heuristic, \topift, which iteratively refines the model by sampling from the current distribution, selecting the lowest-cost samples, and fitting a model to this elite set. Although effective empirically, \topift\ highlights the challenges that motivate the principled algorithm, \alift\ (Section~\ref{sec:alift}).

At iteration $r$, the heuristic draws $m\cdot M$ samples from the current model $\L_{r-1}$, evaluates the zeroth-order cost $d(s)$ for each sample, selects the best $m$ samples, and learns $\L_{r-1}$ using this set to obtain $\L_r$. The cost function $d(\cdot)$ thus \emph{guides} the refinement of the model in conjunction with the prior. 

\begin{algorithm}[htb]
  \caption{\topift: Iterated Fitting of Targets Using Top Samples.\label{alg3}}
  \KwIn{$m, M, Q$}
  \KwData{$\mathcal{L}$ (initial model)}
  \BlankLine
  $\L_0 \leftarrow \L$ \tcp*{initialize}
  \For{$r \leftarrow 1$ \KwTo $Q$}{
    sample $S_r$ from $\L_{r-1}$ with $|S_r| = m\cdot M$ \tcp*{generate samples}
    evaluate $d(s)$ for all $s\in S_r$ \tcp*{compute cost}
    $S'_r \leftarrow$ $m$ samples in $S_r$ with smallest $d(s)$ \tcp*{elite set}
    $\L_r \leftarrow$ fit a model to the samples $S'_r$ \tcp*{update model}
  }
  \Return $\L_Q$
\end{algorithm}

The heuristic in \cref{alg3} is closely related to the cross-entropy (CE) method~\citep{rubinstein1999cross}, where an ``elite'' set of high-performing samples is repeatedly used to update a parametric model via MLE. In our setting, the generative model plays the role of the parametric family, and model-fitting corresponds to the MLE update. However, CE-style updates are difficult to analyze, since the hard-min elite-selection step can induce mode collapse and eliminate coverage of important regions.

These difficulties manifest directly in \topift: 
Its hard-min selection step is analytically brittle, and repeated model-fitting without explicit control over density ratios can progressively erode probability mass on rare but critical regions. \alift\ resolves these issues by replacing hard-min with a controlled annealing schedule and by introducing Metropolis--Hastings corrections (\cref{sec:alift}), ensuring that samples at each iteration follow the correct intermediate target distribution; this prevents coverage errors from compounding across iterations. We present the connection between \alift and \topift in more detail in \cref{app:topift}, and empirically demonstrate the need for iterated distribution correction in \cref{sec:empirical_10d}. 

To implement the Metropolis--Hastings correction and hence show our theoretical results, we require the following assumption. This assumption holds whenever the generative model admits tractable likelihood evaluation, including many autoregressive sequence models and open weight models.

\begin{assume}
\label{ass:llm-prob}
For any $s\in S$, the probability value $\L(s)$ is available. This must hold not only for the initial model but also for any intermediate models obtained.
\end{assume}



\subsection{\alift Algorithm}
\label{sec:alift}

Our final algorithm, \alift, is presented in \cref{alg1}. The input to \alift is the number of samples ($m = \Poly(\maxtemp, D)$), and the parameter $\maxtemp$. We set $M = m^2$. Recall that $D = \max_{s \in S} d(s)$. 

In this algorithm, we intend the model $\L_{\temp}(s)$ to be an approximation to $p_{\temp}(s) \propto \L(s) \cdot e^{-\temp \cdot d(s)}$, where the temperature parameter $\temp$ will gradually increase from $0$ to  $\maxtemp$. The main difference between \alift and \topift is that the probability of retaining a sample is adjusted based on the target distribution $p_{\temp}$.  This adjustment is an interesting application of the Metropolis--Hastings algorithm~\citep{metropolis}, which runs the IMH chain for $M$ steps until mixing. We require \cref{ass:llm-prob} for both $\L$ and the intermediate learned  models $\model_{\temp}$, in particular to enable MCMC sampling from $p_{\temp}$ given a model for $p_{\prevtemp}$. As discussed in \cref{sec:analysis}, this prevents errors in model learning from accumulating across iterations.

\begin{algorithm}[htbp]
  \caption{\alift: Algorithm Driven Iterated Fitting of Targets.}\label{alg1}
  \KwIn{$m, \maxtemp$}
  \KwData{$\mathcal{L}$ (base model), $D$ (temperature step denominator)} 
  \BlankLine
  $\model_{0} \leftarrow \mathcal{L}$\;
  $\temp \leftarrow 0$\;
  $M \leftarrow m^{2}$\tcp*{choice of $M$ justified in \cref{thm:main_iter}.}
  \BlankLine
  \While{$\temp \le \maxtemp$}{
    $\prevtemp \leftarrow \temp$;\;
    $\tau \leftarrow \min\left(\tau + \dfrac{1}{D}, T\right)$;\;
    $S_{\tau} \leftarrow \varnothing$\tcp*{initialize sample set at this temperature}
    \For{$k \leftarrow 1$ \KwTo $m$}{
      Sample $s_{0}\sim \model_{\prevtemp}$;\; $w_{\temp}(s_0) \leftarrow \mathcal{L}(s_0)\cdot e^{-\temp\cdot d(s_0)}$\tcp*{$\prevtemp$ is the previous value of $\temp$}
      \For{$i \leftarrow 1$ \KwTo $M$}{
        Sample $\hat{s}_{i}\sim \model_{\prevtemp}$;\;
        $w_{\temp}(\hat{s}_{i}) \leftarrow \mathcal{L}(\hat{s}_{i})\cdot e^{-\temp\cdot d(\hat{s}_{i})}$\tcp*{weight; note $p_{\temp}(s)\propto w_{\temp}(s)$}
        $\beta_{i} \leftarrow \min\!\left(1,\; \dfrac{w_{\temp}(\hat{s}_{i})}{\model_{\prevtemp}(\hat{s}_{i})}\cdot
          \dfrac{\model_{\prevtemp}(s_{i-1})}{w_{\temp}(s_{i-1})}\right)$\tcp*{MH acceptance probability}
        \eIf{$u\sim\mathrm{Uniform}(0,1)$ satisfies $u \le \beta_{i}$}{
          $s_{i} \leftarrow \hat{s}_{i}$\;
        }{
          $s_{i} \leftarrow s_{i-1}$\;
        }
      }
      $S_{\tau} \leftarrow S_{\tau} \cup \{\, s_{M}\,\}$\tcp*{append final sample into multiset of samples}
    }
    $\model_{\temp} \leftarrow$ fit a model to the samples $S_{\tau}$\tcp*{\label{step8} $\model_{\temp}$ is the learned model for $p_{\temp}$}
  }
  \Return $\model_{\maxtemp}$\;
\end{algorithm}


\noindent {\bf Remarks.} In \cref{alg1}, we use a single complexity parameter $m$ to control multiple aspects of the algorithm for simplicity of analysis. As stated in \cref{ass:learn}, $m$ primarily serves as a learnability parameter, i.e., the number of samples required. We also use $m$ to determine both the number of Metropolis--Hastings iterations (running the step $m$ times) and the length of each iteration ($M = m^2$). This ensures that the computational effort per simulated annealing step scales polynomially with the learnability parameter. Further, in certain cases like the setting in \cref{app:nonconvex_mbo,sec:empirical_10d}, the dependence of the run-time on $D$ can be removed via a geometric annealing schedule. We also note that the model-fitting step is efficient in practice if we use the parameters of $\L_{\tau-}$ as an efficient starting point to learn $\model_\tau$, preventing the need to learn from scratch in each iteration. Finally,  \cref{alg1} is an instance of Model Reference Adaptive Search (MRAS)~\citep{hu2007model}, which generalizes the cross-entropy method~\citep{rubinstein1999cross}. We present this connection in \cref{sec:mras_connection}.

\section{Coarse Learnability and Sample Complexity of \alift} 
\label{sec:coarse}
\label{sec:coarse1}
\label{sec:analysis}
We will analyze the sample complexity of \alift in \cref{sec:analysis}.  To do this,  we need an assumption on the learning agent. In \cref{sec:coarse1}, we propose the novel ``coarse learnability'' assumption, which essentially says that if we can approximately generate samples from $p_{\temp}$, then we can coarsely learn a model $\model_{\temp}$ for it using the samples. Coarse learnability asks only that model fitting produce a proposal with tails heavy enough to cover the target up to polynomial density ratios (in the number of samples used) on all but exponentially rare target mass. 




For two distributions $p$ and $L$ on the same support $S$, we define the
\emph{coverage} of $L$ with respect to $p$ at a point $s \in S$ as
$
\mathrm{cov}_{p,L}(s) \;=\; \frac{p(s)}{L(s)}.
$
This notion, similar to analogous notions in reinforcement learning and inference-time alignment
(e.g.,~\citep{xie2022coverage,huang2025inferencealignment}),
quantifies how well the proposal distribution $L$ ``covers'' the probability mass of the
target $p$: large values indicate under-coverage, while small values indicate
over-coverage.

\begin{assume}[Coarse Learnability]
\label{ass:learn}
For each temperature $\tau \in [0,T]$, let $p_{\temp}(s) \propto \model_0(s) \cdot e^{-\tau d(s)}$ be the target distribution and $\model_{\temp}$ the model obtained after fitting. We say the model family satisfies \emph{coarse learnability} if for any large enough $K \ge \mathrm{poly}(D, T)$ and  any error $\epsilon \in (0,1)$ and confidence $\delta \in (0,1)$, there is a sample size $m = \mathrm{poly}(K/\delta, \ln 1/\epsilon)$ such that the following holds:

If $\model_{\temp}$ is trained using $m$ samples from a distribution $\hat{p}$, 
then with probability at least $1 - \delta$ (over the training samples), the learned model satisfies:
\begin{equation}
\label{eq:fine}
    \Pr_{s\sim p_{\tau}}\!\left[ \mathrm{cov}_{p_{\temp},\model_{\temp}}(s) > K \right] \le  O\left(\tv(\hat{p}, p_{\temp})\right) + \epsilon.
\end{equation}
\end{assume}

The key distinction between \cref{ass:learn} and standard density estimation or PAC learning \citep{Valiant} lies in the error regime. Standard guarantees typically yield error rates polynomial in the sample size ($\epsilon \approx \text{poly}(1/m)$). In contrast, our sample complexity $m$ scales with $\ln(1/\epsilon)$, which means polynomial samples yield \emph{exponentially small} error probabilities ($\epsilon \approx e^{-m}$), albeit for a coarse coverage guarantee. This stronger tail bound is needed for the following reason. Since we assumed the global optimum could have exponentially small probability under the initial prior, a learner with only polynomial error guarantees could validly assign vanishing probability to the optimal region. The exponential tail bound ensures that the intermediate models preserve coverage of these rare regions throughout the annealing schedule.

 To build intuition, we consider the following simplification. Assume $\hat{p} = p_{\tau}$. For given $K$, set $\epsilon = e^{-K}$ and $\delta = 1/K$. \cref{ass:learn} now implies $K = m^{\beta}$ for some constant $\beta \in (0,1)$. If we re-parametrize in terms of $m$, we have $\epsilon = e^{-m^{\beta}}$. In this regime, $K \le m$, so that we can use $m$ instead of $K$ in \cref{eq:fine}, which yields 
$\Pr\left[ \mathrm{cov}_{p_{\temp},\model_{\temp}}(s) > m \right] \le \epsilon = e^{-m^{\beta}}.$   
Then, \cref{ass:learn} simplifies to the following property for general parametric models and targets:
\begin{quote} {\bf \cref{ass:learn}, special case.} A model $\mathcal{L}$ coarsely learns a parametric distribution $p$ if for sufficiently large sample size $m$ that depends polynomially on the parameters of the distribution, w.p. $1-1/m^{\beta}$, the model $\mathcal{L}^m$ trained on $m$ samples from $p$ satisfies  $\Pr\left[\text{cov}_{p,\mathcal{L}^m}(s) > m \right] \le e^{-m^{\beta}}$, where $\beta > 0$ is a small constant.
\end{quote}
Coarse learnability guarantees polynomial coverage of the target density, except on an atypical set of exponentially small measure. Unlike density estimation guarantees that require the total variation distance to vanish, this condition allows for persistent approximation errors (model mis-specification; see \cref{app:miss}). This extends coverage-based analyses~\citep{huang2025inferencealignment} to the iterative setting, where as long as the coverage remains polynomially bounded, the M-H correction can recover the exact target distribution, bridging regimes where the initial coverage gap is exponentially large via intermediate polynomial steps. While this appears to be a strong assumption, such intuition is implicit in classical MBO \citep{rubinstein1999cross, hu2007model}, and is fundamentally linked to statistical estimators that prioritize heavy tails (e.g., over-smoothed KDE). In \cref{sec:empirical_10d}, we empirically demonstrate a non-convex optimization setting where an inflated proposal with distribution correction allow for the global convergence of \alift when sample budgets are small. In \cref{sec:justify}, we provide evidence supporting the plausibility of \cref{ass:learn} for well-known generative distributions. 

Our main analytical result analyzes the sample complexity of \alift under coarse learnability. 

\begin{theorem} [Proved in \cref{app:main_proof}]
\label{thm:main_iter}
Under  \cref{ass:learn}, for sufficiently large $m = \mbox{poly}(T,D)$ 
the sample complexity of \alift is $O(m M T D) = \Poly(T, D)$, and with high probability, the sampling distribution $\hat{p}$ of $S_{\maxtemp}$ satisfies\footnote{Since $p_{\temp}$ is only coarsely learnable via $\model_{\temp}$, we need to consider the $\tv$ with respect to the sampling distribution $S_{\temp}$, which corrects this coarse learning error via the Metropolis--Hastings procedure.} $\tv(\hat{p}, p_{\maxtemp}) = O\left(e^{-m^{\alpha}}\right)$, where $\alpha > 0$ is a constant.
\end{theorem}

\subsection{Proof of \cref{thm:main_iter}} 
\label{app:main_proof}

Our proof focuses on a particular iteration $\tau > 0$. We consider the inner loops of \aldrift, where we generate the set of samples $S_\tau$ using Metropolis-Hastings.
Let $W_{\tau^-} = \{s \mid p_{\tau^-}(s) \le m \cdot \mathcal{L}_{\tau^-}(s)\}$ be the ``typical set'' where the proposal covers the previous target, and let $\eta = \Pr_{p_{\tau^-}}[W_{\tau^-}^c]$ be the mass of the atypical set.

\begin{lemma}
\label{lem:mixing}
At step $\temp$, let $\mathcal{E}$ be the event that the Metropolis-Hastings chain in \alift encounters a state $s \in W_{\tau^-}^c$ at any step during the generation of the $m$ samples in $S_\tau$. 
With probability at least $1 - m^2 \eta$, the event $\mathcal{E}$ does not occur. Conditioned on $\neg \mathcal{E}$, the distribution $\hat{p}$ of any sample generated by the chain satisfies:
\begin{equation}
    \label{eq:delta}
    \tv(\hat{p},p_{\tau}) \le e \cdot \eta +  e^{-m/e}.
\end{equation}
\end{lemma}
\begin{proof} First, since $p_{\tau^-}(s) > m \cdot \mathcal{L}_{\tau^-}(s)$ for all $s \in W_{\prevtemp}^c$, we have
$$\Pr_{s \sim \model_{\prevtemp}} [s \in W_{\prevtemp}^c] \le \frac{1}{m} \Pr_{p_{\prevtemp}}[W_{\prevtemp}^c] = \frac{\eta}{m}.$$
The algorithm takes $M=m^2$ steps per sample, for a total of $m$ samples.  By a union bound over all these steps, the probability that a state in $W_{\prevtemp}^c$ is ever proposed is at most $m^3 \cdot (\eta/m) = m^2 \eta$. This establishes the probability bound for event $\mathcal{E}$.

Conditioned on $\neg \mathcal{E}$, the chain operates entirely within $W_{\prevtemp}$.  Since $p_{\temp}(s) \propto p_{\prevtemp}(s) e^{-d(s)/D}$ and since $d(s) \in [0,D]$ for all $s$, this means $p_{\tau}(s) \le e \cdot p_{\tau^-}(s)$. 
Thus, for any $s \in W_{\prevtemp}$, the density ratio with respect to the \emph{new} target is bounded:
$$ \frac{p_{\temp}(s)}{\model_{\prevtemp}(s)} \le e \cdot \frac{p_{\prevtemp}(s)}{\model_{\prevtemp}(s)} \le e \cdot m. $$
The sampling procedure is therefore equivalent to an independent Metropolis--Hastings (IMH) chain restricted to $W_{\prevtemp}$ with density ratio bounded by $em$. By the uniform ergodicity of IMH~\citep{mengersen1996rates,Tierney}, the variation distance to the restricted target $p'_{\tau} \propto p_\tau|_{W_{\tau^-}}$ after $M=m^2$ steps is:
\[ \tv(\hat{p}, p'_{\tau}) \le \left(1 - \frac{1}{e m}\right)^M \le e^{-m/e}. \]

We finally bound the distance to the true target $p_\tau$. The error introduced by restricting the target to $W_{\prevtemp}$ is bounded by the mass of the excluded set under $p_\tau$:
$$ \tv(p'_\tau, p_\tau) = p_\tau(W_{\prevtemp}^c) \le e \cdot p_{\prevtemp}(W_{\prevtemp}^c) = e \cdot \eta. $$
By the triangle inequality:
$$ \tv(\hat{p}, p_{\tau}) \le \tv(\hat{p}, p'_{\tau}) + \tv(p'_{\tau}, p_{\tau}) \le e^{-m/e} + e \cdot \eta. $$
This completes the proof.
\end{proof}

Using the above lemma, we now complete the proof of \cref{thm:main_iter}.

\begin{proof}[Proof of \cref{thm:main_iter}]
 As in the text following \cref{ass:learn}, we set $\epsilon = \exp(-K)$ and $\delta = 1/K$, so that  $K = m^{\beta}$ for constant $\beta > 0$.  We now choose $m$ large enough that $m^\beta = \omega(TD)$, and note that $m \ge K$.

\newcommand{\prev}{\mathrm{prev}}

We prove by induction that if the algorithm succeeds (does not hit atypical sets), then for all $\tau$, $\tv(\hat{p}, p_{\tau}) = O(e^{-m^\alpha})$ for some constant $\alpha \in (0,\beta)$.  For the base case, note that $\tv(\hat{p}, p_{0}) = 0$ since we begin with the model $\L_0 = \L$. For the inductive step, assume $\tv(\hat{p}, p_{\tau^-}) \le \epsilon_{\prev}$. By \cref{ass:learn}, the learned model $\mathcal{L}_{\tau^-}$ has an atypical set mass $\eta \le O(\epsilon_{\prev}) + e^{-m^\beta}$.
Applying Lemma~\ref{lem:mixing}, the error at the next step is:
$$ \tv(\hat{p}, p_\tau) \le e \cdot \left(O\left(\epsilon_{\prev}\right) + e^{-m^\beta}\right) + e^{-m/e}. $$
Unrolling this recurrence over $T \cdot D$ steps results in an error growth of $\phi^{TD}$ for some constant $\phi > 1$. Thus:
$$ \tv(\hat{p}, p_T) = O\left(\phi^{TD} \cdot e^{-m^\beta} \right) = e^{O(TD) - m^\beta}. $$
Assuming $m^\beta = \omega(TD)$, we have $\tv(\hat{p}, p_T) = O(e^{-m^\alpha})$. This completes the induction.
 
The algorithm fails at any step $\temp$ if (a) learning fails (probability $\delta$) or (b) sampling hits $W^c$ (probability $m^2 \cdot \eta$).  The above derivation yields $\eta = O(e^{-m^\alpha})$, so that we have $m^2 \eta = o\left(m^{-\beta}\right)$.  Further, $\delta = 1/K = m^{-\beta}$.
By a union bound over $T \cdot D$ steps, total failure probability is $O(T \cdot D \cdot m^{-\beta})  = o(1)$ since we assume $m^\beta = \omega(TD)$.

Finally, at given any $\temp$, the total number of samples is $O(m^3) = \Poly(\maxtemp, D)$. This completes the proof of \cref{thm:main_iter}.
\end{proof}

\section{Sample Complexity of \alift for Bounded Non-Convex Optimization}
\label{app:nonconvex_mbo}

Classical zeroth-order global optimization in $\mathbb{R}^n$ often relies on local Markov Chain Monte Carlo (MCMC) methods~\citep{KalaiV}. While these methods achieve polynomial time complexity on strictly log-concave target distributions, they can require exponential time to mix on non-convex landscapes due to local energy barriers~\citep{holley1987logarithmic,hajek1988cooling}. We prove that Model-Based Optimization (MBO) overcomes this limitation, and achieves logarithmic sample complexity in the precision parameter, hence providing a finite-sample analog to the asymptotic convergence results for MRAS. 

By leveraging a modification of \alift with an inflated proposal distribution $\model_{\tau}$, we show that it can efficiently (globally) optimize non-convex functions that are bounded by a quadratic envelope. In particular, we show that  \cref{ass:learn} holds at all intermediate steps with a bound $K$ that is independent of the temperature parameter $\tau$, guaranteeing  sample complexity $\widetilde{O}(\log 1/\epsilon)$ to achieve error $\epsilon$ in the global optimum, matching the bounds of space partitioning methods~\citep{munos2011optimistic,bubeck2011x} for this setting (under some caveats that we mention in \cref{app:compare_mbo}). The $O(\cdot)$ notation hides constants that depend on $n,K$, and $\log \frac{1}{\delta}$. 

Recall that in \cref{sec:empirical_10d}, we presented an empirical demonstration of the need for an inflated proposal distribution (with sample correction) in ensuring the convergence of \alift with limited sample budgets, comparing it to \topift that uses the empirical variance and samples without distribution correction, and gets trapped in local optima.

\subsection{Optimization Setting and Algorithm}
The domain is $\mathbb{R}^n$. Assume the generative prior is a Gaussian $\mathcal{L}_0(x) \propto \exp(-U(x))$, where $U(x) = \frac{\mu_0}{2}\|x - x_0\|^2$ for some $\mu_0 > 0$.

Assume the objective $d(x)$ has a unique global minimum at $x^*$.  We assume $d(x)$ is  bounded by a quadratic envelope around its true optimum: there exist constants $\mu_d, L_d > 0$ such that for all $x \in \mathbb{R}^n$,
\begin{align*}
    \frac{\mu_d}{2}\|x - x^*\|^2 \le d(x) - d(x^*) \le \frac{L_d}{2}\|x - x^*\|^2.
\end{align*}
Note that $d(x)$ is not assumed to be convex and may contain arbitrary local non-convexities within these bounds. This model is well-studied in non-convex optimization; see for instance~\citep{munos2011optimistic,bubeck2011x}. 

We now run \alift for $N$ steps as described in \cref{app:exec_alift}. There are changes needed to the algorithm to handle the lack of an upper bound on $d(x)$, and to make the sampling process exact, that we detail there. We also detail the parameter settings there, and re-do the proof of sample complexity.

When the temperature parameter is $\tau$, the target distribution is $p_\tau(x) \propto \exp(-f_\tau(x))$, where $f_\tau(x) = U(x) + \tau d(x)$. The algorithm draws $m$ samples from $p_\tau$ (where the parameter $m$ will be chosen later), computes the empirical mean $\hat{\omega}_\tau$, and fits the proposal distribution (or generative model) $\mathcal{L}_\tau(x) = \mathcal{N}(\hat{\omega}_\tau, \frac{2}{\mu_\tau}I_n)$, where $\mu_\tau = \mu_0 + \tau \mu_d$. We assume the sampling is exactly from $p_{\tau}$, and we justify this in \cref{app:exec_alift}.  It is also crucial for our proof of \cref{thm:funnel_mbo} below that the distribution has larger variance than desired and is set based on the convex envelope parameters $\mu_d, L_d$, hence deviating from standard MBO that uses the sample variance directly. 


\subsection{Verifying  Coarse Learnability}
We now show that  \cref{ass:learn} holds at all intermediate steps of the algorithm, for the proposal distribution $\mathcal{L}_\tau(x) = \mathcal{N}(\hat{\omega}_\tau, \frac{2}{\mu_\tau}I_n)$. Note that in this setting, \cref{ass:learn} is not an assumption; we show that it holds because of our choice of the model $\mathcal{L}_\tau(x)$. 

To simplify presentation, we define the following global geometric constants:
\begin{itemize}
    \item The proxy condition number: $\kappa_{\max} = \max\left(2, \frac{L_d}{\mu_d} \right)$.
    \item The initial shift penalty: $E_{\text{shift}} = \frac{\mu_0}{2}\|x_0 - x^*\|^2$. 
    \item The inflation factor: $B = \kappa_{\max}^{n/2} \exp(E_{\text{shift}})$.
\end{itemize}

We will further assume in the rest of the section that $d(x^*) = 0$, so that $d(x) \ge 0$ for all $x$. This is w.l.o.g., since the normalized density $p_{\tau}$ remains unchanged if we replace $d(x)$ with $\tilde{d}(x) = d(x) - d(x^*)$, so that the algorithm's execution remains identical.

\begin{theorem}[Coarse Learnability of $p_{\tau}$] 
\label{thm:funnel_mbo}
Given the setting and algorithm defined above, if the per-step sample size satisfies $m \ge C \left(\frac{n}{2} \ln \kappa_{\max} + E_{\text{shift}}\right)\left(n + \ln \frac{N}{\delta}\right)$ for constant $C > 0$, then for any temperature $\tau$, the model $\model_{\tau}$ globally satisfies the following condition\footnote{This condition is slightly different from how \cref{ass:learn} is phrased, but suffices for the sample complexity proof in \cref{app:exec_alift}.} with probability at least $1 - \delta/(2N)$:
\begin{align*}
     \sup_{x \in \mathbb{R}^n} \frac{p_\tau(x)}{\mathcal{L}_\tau(x)} \le 2^{n/2} \cdot B^5 \cdot \exp\left(  1 + 2n \right) := K.
\end{align*}
\end{theorem}

\paragraph{Remarks.}   \cref{thm:funnel_mbo} can be viewed as a  special case of the Coarse Learnability assumption (\cref{ass:learn}). Because \cref{thm:funnel_mbo} establishes a global supremum bound on the density ratio, we have $\Pr[\text{ratio} > K] = 0$. Furthermore, because we utilize Rejection Sampling (\cref{app:exec_alift}), the empirical samples are drawn perfectly from the target, meaning $\tv(\hat{p}, p_{\temp}) = 0$. Consequently, the inequality in \cref{ass:learn} is satisfied for any $\epsilon > 0$. Further, the sample complexity is $m = \mbox{poly}\left(\ln K, \ln 1/\delta \right)$, which is stronger than the bound in \cref{ass:learn}, assuming $K$ is a large enough function of the problem parameters (but independent of $T$). \cref{thm:funnel_mbo} suffices for the specialized analysis of \alift that we perform in \cref{app:exec_alift}.

Note  that $K$ (and consequently the algorithm's sample complexity) does not grow with $\tau$. This non-asymptotically captures the limiting behavior predicted by Laplace's method \citep{wong_asymptotic_1989} and the Bernstein-von Mises theorem \citep{van_der_vaart_asymptotic_1998} in Bayesian inference: as the $\tau$-scaled regularizer overwhelms the prior, the prior collapses into a constant spatial penalty evaluated at the optimum. However, while those classical theorems are valid only as $\tau \to \infty$, our quadratic envelope technique provides an upper bound for any finite $\tau \ge 0$.

The rest of this subsection is devoted to proving this theorem.  



\paragraph{Upper Bounding $p_{\tau}$.} We now show a sequence of lemmas that will bound the density ratio. We start with $p_{\tau}$, and bound it via standard methods.

\begin{lemma}[Target Envelope] \label{lem:target_envelope}
The target distribution $p_\tau(x)$ is globally upper-bounded by a scaled Gaussian envelope:
\begin{align*}
    p_\tau(x) \le B \cdot \left(\frac{\mu_\tau}{2\pi}\right)^{n/2} \exp\left( -\frac{\mu_\tau}{2}\|x - \bar{x}_\tau\|^2 \right)
\end{align*}
where $\bar{x}_\tau = \frac{\mu_0 x_0 + \tau \mu_d x^*}{\mu_\tau}$. 
\end{lemma}
\begin{proof}
Let the target probability distribution be defined as $p_\tau(x) = \frac{1}{Z_\tau} \exp(-f_\tau(x))$, where $f_\tau(x) = \frac{\mu_0}{2}\|x - x_0\|^2 + \tau d(x)$ and $Z_\tau = \int_{\mathbb{R}^n} \exp(-f_\tau(x)) dx$ is the normalizing constant. 

By assumption, $d(x)$ is bounded by quadratics: $\frac{\mu_d}{2}\|x - x^*\|^2 \le d(x) \le \frac{L_d}{2}\|x - x^*\|^2$. We can therefore sandwich $f_\tau(x)$ between a lower bounding quadratic $f_{\tau, \text{low}}(x)$ and an upper bounding quadratic $f_{\tau, \text{up}}(x)$:
\begin{align*}
    \underbrace{\frac{\mu_0}{2}\|x - x_0\|^2 + \frac{\tau \mu_d}{2}\|x - x^*\|^2}_{f_{\tau, \text{low}}(x)} \le f_\tau(x) \le \underbrace{\frac{\mu_0}{2}\|x - x_0\|^2 + \frac{\tau L_d}{2}\|x - x^*\|^2}_{f_{\tau, \text{up}}(x)}.
\end{align*}

We first simplify $f_{\tau, \text{low}}(x)$ as follows:
\begin{align*}
    f_{\tau, \text{low}}(x) 
    &= \frac{1}{2}(\mu_0 + \tau \mu_d)\|x\|^2 - (\mu_0 x_0 + \tau \mu_d x^*)^T x + \frac{1}{2}(\mu_0 \|x_0\|^2 + \tau \mu_d \|x^*\|^2).
\end{align*}
To consolidate this into a single quadratic, we define $\mu_\tau = \mu_0 + \tau \mu_d$ and the weighted center $\bar{x}_\tau = \frac{\mu_0 x_0 + \tau \mu_d x^*}{\mu_\tau}$. 
Substituting these into the expansion yields:
\begin{align*}
    f_{\tau, \text{low}}(x) &= \frac{\mu_\tau}{2}\|x\|^2 - \mu_\tau \bar{x}_\tau^T x + \frac{1}{2}(\mu_0 \|x_0\|^2 + \tau \mu_d \|x^*\|^2) \\
 &= \frac{\mu_\tau}{2} \left( \|x\|^2 - 2\bar{x}_\tau^T x + \|\bar{x}_\tau\|^2 \right) - \frac{\mu_\tau}{2}\|\bar{x}_\tau\|^2 + \frac{1}{2}(\mu_0 \|x_0\|^2 + \tau \mu_d \|x^*\|^2) \\
    &= \frac{\mu_\tau}{2}\|x - \bar{x}_\tau\|^2 + C_{\tau, \text{low}},
\end{align*}
where  
\begin{align*}
    C_{\tau, \text{low}} &=  \frac{1}{2} (\mu_0 \|x_0\|^2 + \tau \mu_d \|x^*\|^2 - \mu_\tau \|\bar{x}_\tau\|^2)\\
    &= \frac{1}{2} \left[ \mu_0 \|x_0\|^2 + \tau \mu_d \|x^*\|^2 - \frac{1}{\mu_0 + \tau \mu_d} \|\mu_0 x_0 + \tau \mu_d x^*\|^2 \right] \\
    &= \frac{1}{2(\mu_0 + \tau \mu_d)} \Big[ \mu_0^2\|x_0\|^2 + \mu_0 \tau \mu_d \|x_0\|^2 + \tau \mu_d \mu_0 \|x^*\|^2 + (\tau \mu_d)^2\|x^*\|^2 \\
    &\hspace{3cm} - \mu_0^2 \|x_0\|^2 - 2\mu_0 \tau \mu_d \langle x_0, x^* \rangle - (\tau \mu_d)^2 \|x^*\|^2 \Big] \\ 
    &= \frac{1}{2(\mu_0 + \tau \mu_d)} \Big[ \mu_0 \tau \mu_d \left( \|x_0\|^2 - 2\langle x_0, x^* \rangle + \|x^*\|^2 \right) \Big] \\
    &= \frac{1}{2} \frac{\mu_0 \tau \mu_d}{\mu_0 + \tau \mu_d} \|x_0 - x^*\|^2.
\end{align*}

Analogously, substituting $\mu_d$ with $L_d$, completing the square for $f_{\tau, \text{up}}(x)$ yields a quadratic with smoothness $L_\tau = \mu_0 + \tau L_d$, centered at $\tilde{x}_\tau = \frac{\mu_0 x_0 + \tau L_d x^*}{\mu_0 + \tau L_d}$: 
\begin{align*}
    f_{\tau, \text{up}}(x) = \frac{L_\tau}{2}\|x - \tilde{x}_\tau\|^2 + C_{\tau, \text{up}}, \quad \text{where} \quad C_{\tau, \text{up}} = \frac{1}{2} \frac{\mu_0 \tau L_d}{\mu_0 + \tau L_d} \|x_0 - x^*\|^2.
\end{align*}

Observe now that the fractional term in $C_{\tau, \text{up}}$:
\begin{align*}
    \frac{\tau L_d}{\mu_0 + \tau L_d} = 1 - \frac{\mu_0}{\mu_0 + \tau L_d} \le 1.
\end{align*}
 Thus, for all $\tau \ge 0$:
\begin{align*}
    C_{\tau, \text{up}} \le \frac{\mu_0}{2} \|x_0 - x^*\|^2 = E_{\text{shift}}.
\end{align*}
Since $C_{\tau, \text{low}} \ge 0$, we have:
\begin{align*}
    C_{\tau, \text{up}} - C_{\tau, \text{low}} \le C_{\tau, \text{up}} \le E_{\text{shift}}.
\end{align*}

Since $f_\tau(x) \le f_{\tau, \text{up}}(x)$ for all $x$, it follows that $\exp(-f_\tau(x)) \ge \exp(-f_{\tau, \text{up}}(x))$. We now lower-bound the normalizer $Z_\tau$ as:
\begin{align*}
    Z_\tau = \int_{\mathbb{R}^n} \exp(-f_\tau(x)) dx &\ge \int_{\mathbb{R}^n} \exp(-f_{\tau, \text{up}}(x)) dx \\
    &= \int_{\mathbb{R}^n} \exp\left( -\frac{L_\tau}{2}\|x - \tilde{x}_\tau\|^2 - C_{\tau, \text{up}} \right) dx \\
    &= e^{-C_{\tau, \text{up}}} \int_{\mathbb{R}^n} \exp\left( -\frac{L_\tau}{2}\|x - \tilde{x}_\tau\|^2 \right) dx.
\end{align*}

The remaining integral is an unnormalized multivariate Gaussian with covariance matrix $\Sigma = \frac{1}{L_\tau} I$. Since $\int \exp(-\frac{1}{2}x^T \Sigma^{-1} x) dx = \sqrt{(2\pi)^n \det(\Sigma)}$, we get:
\begin{align*}
    Z_\tau \ge e^{-C_{\tau, \text{up}}} \left(\frac{2\pi}{L_\tau}\right)^{n/2}.
\end{align*}

We finally upper-bound the numerator of $p_\tau(x)$ as: $f_\tau(x) \ge f_{\tau, \text{low}}(x) \implies \exp(-f_\tau(x)) \le \exp(-f_{\tau, \text{low}}(x))$. Applying this to the numerator and dividing by our lower bound for $Z_\tau$:
\begin{align*}
    p_\tau(x) = \frac{\exp(-f_\tau(x))}{Z_\tau} &\le \frac{\exp\left( -\frac{\mu_\tau}{2}\|x - \bar{x}_\tau\|^2 - C_{\tau, \text{low}} \right)}{e^{-C_{\tau, \text{up}}} \left(\frac{2\pi}{L_\tau}\right)^{n/2}} \\
    &= \left(\frac{L_\tau}{2\pi}\right)^{n/2} \exp(C_{\tau, \text{up}} - C_{\tau, \text{low}}) \exp\left(-\frac{\mu_\tau}{2}\|x - \bar{x}_\tau\|^2 \right) \\
    & = \left(\frac{L_\tau}{\mu_\tau}\right)^{n/2} e^{E_{\text{shift}}} \left(\frac{\mu_\tau}{2\pi}\right)^{n/2} \exp\left(-\frac{\mu_\tau}{2}\|x - \bar{x}_\tau\|^2 \right).
\end{align*}
Since $L_\tau / \mu_\tau \le \kappa_{\max}$, we can bundle the leading factors into a single constant $B = \kappa_{\max}^{n/2} e^{E_{\text{shift}}}$, yielding the final bound:
\begin{align*}
    p_\tau(x) \le B \cdot \left(\frac{\mu_\tau}{2\pi}\right)^{n/2} \exp\left( -\frac{\mu_\tau}{2}\|x - \bar{x}_\tau\|^2 \right).
\end{align*}
This completes the proof.
\end{proof}

The previous lemma immediately yields the following corollary:

\begin{corollary}\label{cor:normalize}
The normalizer $Z_\tau$ is bounded as:
$$ Z^{\text{low}}_\tau := e^{-C_{\tau, \text{up}}} \left(\frac{2\pi}{L_\tau}\right)^{n/2} \le Z_\tau \le e^{-C_{\tau, \text{low}}} \left(\frac{2\pi}{\mu_\tau}\right)^{n/2} := Z^{\text{up}}_\tau.$$
In particular, the ratio between these bounds is:
$$ \frac{Z^{\text{up}}_\tau}{Z^{\text{low}}_\tau} \le e^{E_{\text{shift}}} \cdot \left(\frac{L_{\tau}}{\mu_{\tau}}\right)^{n/2} \le e^{E_{\text{shift}}} (\kappa_{\max})^{n/2} = B. $$
\end{corollary}

\paragraph{Sub-Gaussian Upper Bound.} Let $q(x) = \mathcal{N}(\bar{x}_\tau, \frac{1}{\mu_\tau}I_n)$. From \cref{lem:target_envelope}, we have: $p_\tau(x) \le B \cdot q(x)$. We now have the following lemmas using this inequality, which will help us bound the density ratio. We provide the proof of the lemma below for completeness.


\begin{lemma} \label{lem:asym}
Let $\tilde{\omega}_\tau = \mathbb{E}_{p_\tau}[x]$ be the true expectation. Then,
$$\mu_\tau\|\tilde{\omega}_\tau - \bar{x}_\tau\|^2 \le \left(\sqrt{n} + \sqrt{2 \ln B}\right)^2 \le 2n + 4 \ln B.$$
\end{lemma}
\begin{proof}
By Jensen's inequality, $\|\tilde{\omega}_\tau - \bar{x}_\tau\|^2 \le \mathbb{E}_{p_\tau}[\|x - \bar{x}_\tau\|^2]$. Let $z = \sqrt{\mu_\tau} (x - \bar{x}_\tau)$. Under the proposal distribution $q$, $z \sim \mathcal{N}(0, I_n)$, which we denote by $\phi(z)$. Under the target distribution, $z$ follows some density $p(z)$. Since $z$ and $x$ are linearly related, we have $p(z) \le B \phi(z)$. We seek to bound $\mathbb{E}_p[\|z\|^2]$. Since $p(z) \le B \phi(z)$, the Kullback-Leibler divergence is  bounded:
$$ D_{KL}(p \| \phi) = \mathbb{E}_p\left[\ln \frac{p(z)}{\phi(z)}\right] \le \mathbb{E}_p[\ln B] = \ln B. $$

By the Donsker-Varadhan  representation~\cite{boucheron}, for any measurable function $f(z)$ and any $\lambda > 0$:
$$ \mathbb{E}_p[\lambda f(z)] - \ln \mathbb{E}_\phi[e^{\lambda f(z)}] \le D_{KL}(p \| \phi) \implies \mathbb{E}_p[f(z)] \le \frac{1}{\lambda} \left( D_{KL}(p \| \phi) + \ln \mathbb{E}_\phi[e^{\lambda f(z)}] \right). $$

We choose $f(z) = \|z\|^2$. Under $\phi(z)$, $\|z\|^2$ follows a $\chi^2_n$ distribution, whose moment generating function is $\mathbb{E}_\phi[e^{\lambda \|z\|^2}] = (1 - 2\lambda)^{-n/2}$ for $\lambda < 1/2$. Substituting this and our KL bound yields:
$$ \mathbb{E}_p[\|z\|^2] \le \frac{\ln B}{\lambda} - \frac{n}{2\lambda} \ln(1 - 2\lambda). $$

Using $-\ln(1 - x) \le \frac{x}{1 - x}$ for $x \in (0, 1)$, we set $x = 2\lambda$ to obtain:
$$ \mathbb{E}_p[\|z\|^2] \le \frac{\ln B}{\lambda} + \frac{n}{1 - 2\lambda}. $$

Choosing $\lambda = \left(2 + \sqrt{\frac{2n}{\ln B}}\right)^{-1} \in (0, 1/2)$  yields:
$$ \mathbb{E}_p[\|z\|^2] \le \left(\sqrt{n} + \sqrt{2 \ln B}\right)^2 \le 2n + 4\ln B. $$
Replacing $z$ with $\sqrt{\mu_\tau} (x - \bar{x}_\tau)$ concludes the proof.
\end{proof}

Before proceeding, we formally define the \emph{variance proxy} of a sub-Gaussian random variable~\citep{vershynin_high-dimensional_2018}. A random variable $Z$ with mean $\mu_Z$ is sub-Gaussian with variance proxy $\nu^2$  if $Z$ satisfies the tail bound $\Pr(|Z-\mu_Z| \ge t) \le 2\exp(-t^2/\nu^2)$ for all $t \ge 0$. A random vector $X \in \mathbb{R}^n$ is sub-Gaussian with variance proxy $\nu^2$ if its one-dimensional projections $\langle X - \mu_X, u \rangle$ onto any unit vector $u \in \mathbb{S}^{n-1}$ are sub-Gaussian with variance proxy $\nu^2$. The following lemma is now straightforward (see~\citep{vershynin_high-dimensional_2018}).

\begin{lemma} \label{lem:subgauss}
 $X \sim p_\tau$ is sub-Gaussian with a variance proxy bounded by $\nu^2 = O\left( \frac{1 + \ln B}{\mu_\tau} \right)$.
\end{lemma}

We will need the following concentration bound; see~\citep{vershynin_high-dimensional_2018} for a proof.

\begin{lemma}[Empirical Mean of Sub-Gaussian Vectors] \label{lem:subgaussian_mean}
Let $X_1, \dots, X_m$ be independent, identically distributed random vectors in $\mathbb{R}^n$ with mean $\omega = \mathbb{E}[X_1]$. Assume each $X_i$ is sub-Gaussian with variance proxy $\nu^2$. Let $\hat{\omega} = \frac{1}{m}\sum_{i=1}^m X_i$ be the empirical mean.  Then, there exists a  constant $C > 0$ such that for any $t > 0$, with probability at least $1 - e^{-t}$:
\begin{align*}
    \|\hat{\omega} - \omega\|^2 \le \frac{C \nu^2}{m} (n + t).
\end{align*}
\end{lemma}

\paragraph{Proof of \cref{thm:funnel_mbo}.} Using these ingredients, we now prove \cref{thm:funnel_mbo}.

\begin{proof}[Proof of \cref{thm:funnel_mbo}]
We evaluate the density ratio $r_\tau(x) = p_\tau(x) / \mathcal{L}_\tau(x)$. The proposal generative model is $\mathcal{L}_\tau(x) = \mathcal{N}(\hat{\omega}_\tau, \frac{2}{\mu_\tau}I_n)$ with density $(\frac{\mu_\tau}{4\pi})^{n/2} \exp(-\frac{\mu_\tau}{4}\|x - \hat{\omega}_\tau\|^2)$. Using the  upper bound from \cref{lem:target_envelope} for the numerator:
\begin{align*}
    r_\tau(x) \le \frac{  B \left(\frac{\mu_\tau}{2\pi}\right)^{n/2} \exp\left( -\frac{\mu_\tau}{2}\|x - \bar{x}_\tau\|^2 \right) }{ \left(\frac{\mu_\tau}{4\pi}\right)^{n/2} \exp\left(-\frac{\mu_\tau}{4}\|x - \hat{\omega}_\tau\|^2\right) } = B 2^{n/2} \exp\left( -\frac{\mu_\tau}{2}\|x - \bar{x}_\tau\|^2 + \frac{\mu_\tau}{4}\|x - \hat{\omega}_\tau\|^2 \right).
\end{align*}
We apply triangle inequality: $\|x - \hat{\omega}_\tau\|^2 \le 2\|x - \bar{x}_\tau\|^2 + 2\|\hat{\omega}_\tau - \bar{x}_\tau\|^2$ to the second term in the exponent.  The $x$-dependent terms  cancel and we have:
\begin{align}
    r_\tau(x) \le 2^{n/2} \cdot B \cdot \exp\left( \frac{\mu_\tau}{2}\|\hat{\omega}_\tau - \bar{x}_\tau\|^2 \right). \label{eq:ratio_bound_funnel}
\end{align}

 By the triangle inequality, $\|\hat{\omega}_\tau - \bar{x}_\tau\|^2 \le 2\|\hat{\omega}_\tau - \tilde{\omega}_\tau\|^2 + 2\|\tilde{\omega}_\tau - \bar{x}_\tau\|^2$. The exponent in the above expression is therefore bounded by $\mu_\tau\|\hat{\omega}_\tau - \tilde{\omega}_\tau\|^2 + \mu_\tau\|\tilde{\omega}_\tau - \bar{x}_\tau\|^2$. 
From \cref{lem:asym}, we have $\mu_\tau\|\tilde{\omega}_\tau - \bar{x}_\tau\|^2 \le 2n + 4 \ln B$. 

To bound $\mu_\tau\|\hat{\omega}_\tau - \tilde{\omega}_\tau\|^2$, note that $\hat{\omega}_\tau$ is  the empirical mean of $m$ i.i.d. draws from the target distribution $p_\tau$, while $\tilde{\omega}_\tau$ is the true mean.  By \cref{lem:subgauss}, these independent draws from $p_\tau$ are sub-Gaussian random vectors with a variance proxy bounded by:
\begin{align*}
    \nu^2 = C_{\text{sg}} \frac{1 + \ln B}{\mu_\tau}
\end{align*}
for some constant $C_{\text{sg}} > 0$. We now apply \cref{lem:subgaussian_mean}. Setting the failure probability to $e^{-t} = \frac{\delta}{2N}$, which implies $t = \ln\left(\frac{2N}{\delta}\right)$, we obtain with  probability at least $1 - \frac{\delta}{2N}$:
\begin{align*}
    \|\hat{\omega}_\tau - \tilde{\omega}_\tau\|^2 & \le \frac{C}{m} \nu^2 \left( n + \ln\left(\frac{2N}{\delta}\right) \right) \\
    & = \frac{C}{m} \left( C_{\text{sg}} \frac{1 + \ln B}{\mu_\tau} \right) \left( n + \ln\left(\frac{2N}{\delta}\right) \right).
\end{align*}
so that
\begin{align*}
    \mu_\tau \|\hat{\omega}_\tau - \tilde{\omega}_\tau\|^2 \le \frac{C \cdot C_{\text{sg}}}{m} (1 + \ln B) \left( n + \ln\left(\frac{2N}{\delta}\right) \right).
\end{align*}

To satisfy our requirement that $\mu_\tau\|\hat{\omega}_\tau - \tilde{\omega}_\tau\|^2 \le 1$, we set:
\begin{align*}
    m \ge C' (1 + \ln B) \left( n + \ln\left(\frac{2N}{\delta}\right) \right)
\end{align*}
where $C' = C \cdot C_{\text{sg}}$. Therefore with probability at least $1 - \frac{\delta}{2N}$, we have $\frac{\mu_\tau}{2}\|\hat{\omega}_\tau - \bar{x}_\tau\|^2 \le 1 + 2n + 4 \ln B$. Substituting this into \cref{eq:ratio_bound_funnel} yields the bound on $K$.
\end{proof}

\subsection{Implementing \alift via Geometric Annealing and Rejection Sampling} 
\label{app:exec_alift}
Given that we have a simple analytic generative model, we make two changes that both simplify \alift and lead to an exponentially improved sample complexity. Further note that we did not assume an upper bound on $d(x)$ in this setting, which needs a different approach than \cref{thm:main_iter}.

In the description below, we set the number of generated samples $m$ for each temperature from the statement of \cref{thm:funnel_mbo}.

\begin{description}
\item[Geometric Annealing.] We use a {\em geometric schedule} of temperatures $\tau$. In particular, we set $\tau_0 = 0$, $\tau_1 = \frac{\mu_0}{L_d}$, and $\tau_{k+1} = (1+\gamma) \cdot \tau_{k}$ for $k \ge 1$, where $\gamma = 1/n$. Since $\tau_N \ge T$, this implies the number of outer iterations of \alift (temperature settings) is $N = O\left(n \ln \frac{T L_d}{\mu_0}\right)$. Such a geometric schedule not only improves the dependence on $T$ exponentially, but is also robust to $d(x)$ being unbounded.
\item[Rejection Sampling.] In the inner loop, to generate $m$ samples from $p_{\tau_{k+1}}$, we use rejection sampling instead of Metropolis-Hastings. Although the normalizing constant $Z_{\tau_{k+1}}$ is unknown, we use $\bar Z_\tau := \left(2\pi/\mu_\tau\right)^{n/2}$ and define the sub-normalized density 
$$\hat p_\tau(x) := \frac{\exp(-f_\tau(x))}{\bar Z_\tau}.$$ By \cref{cor:normalize}, $Z_\tau \le \bar Z_\tau$.  At each sampling step, we generate $x \sim \model_{\tau_k}$ and accept it with probability $$\Pr[\mbox{Accept}\mid x] = \frac{1}{\sqrt{e} K B} \frac{\hat{p}_{\tau_{k+1}}(x)}{\model_{\tau_k}(x)}.$$ 
To generate a sample, we run this process for at most $t_s = 10 K B^2 \ln \frac{2Nm}{\delta}$ steps, declaring failure if a sample is not generated. We generate $m$ samples and use them to fit $\model_{\tau_{k+1}}$.
\end{description}

\paragraph{Remark.} This implementation does not require knowing the exact normalizing constant $Z_\tau$ in order to evaluate the acceptance probability. It only requires a valid upper bound on the envelope constant $B$. If a loose bound $R \ge \|x_0-x^*\|^2$ is known, we may replace $B$ throughout by the computable upper bound $\widehat B := \kappa_{\max}^{n/2}\exp(\mu_0 R/2)$. Under the normalization $d(x^*)=0$, the lower quadratic envelope also gives the computable choice $R=2d(x_0)/\mu_d$.

Focusing just on $T$ and ignoring the problem-dependent parameters of $B, n, K$, the total sample complexity is therefore $\tilde{O}\left(\ln T \cdot m\right) = \widetilde{O}\left(\ln T \cdot \ln \frac{N}{\delta} \right) = \widetilde{O}\left(\ln T \cdot \ln \frac{\ln T}{\delta} \right)$, and we show below that the algorithm fails with probability at most $\delta$.

\subsubsection{Analysis of the Sampling Step}
We first show that the rejection sampling step is feasible, meaning that the accept probability is at most one. Clearly, if this is true, this process will generate $m$ unbiased samples from $p_{\tau_{k+1}}$.

\begin{lemma} \label{lem:geometric_ratio}
Let $\tau' := \tau_{k+1}$ and $\tau = \tau_k$ and let $\gamma$ be the multiplicative growth rate of the annealing procedure. Then, for all iterations $k \ge 0$, with probability $1 - \frac{\delta}{2N}$, we have:
\begin{align*}
    \frac{p_{\tau'}(x)}{\mathcal{L}_\tau(x)} \le K \cdot B \cdot (1+\gamma)^{n/2}.
\end{align*}
Consequently, setting the geometric rate to $\gamma = \frac{1}{n}$ yields $\sup_x \frac{p_{\tau'}(x)}{\model_\tau(x)} \le B\cdot K \cdot \sqrt{e}$.
\end{lemma}
\begin{proof}
We decompose the density ratio into two terms:
\begin{align*}
    \frac{p_{\tau'}(x)}{\mathcal{L}_\tau(x)} = \frac{p_{\tau'}(x)}{p_\tau(x)} \cdot \frac{p_\tau(x)}{\mathcal{L}_\tau(x)} := r(x) \cdot \frac{p_{\tau'}(x)}{p_\tau(x)}.
\end{align*}
From \cref{thm:funnel_mbo}, we already know that  with probability $1-\frac{\delta}{2N}$, we have $r(x) \le K$ for all $x$. 

To bound the first term, we expand the target densities:
\begin{align*}
    \frac{p_{\tau'}(x)}{p_\tau(x)} = \frac{\frac{1}{Z_{\tau'}} \exp(-f_{\tau'}(x))}{\frac{1}{Z_\tau} \exp(-f_\tau(x))} = \frac{Z_\tau}{Z_{\tau'}} \exp\Big( -(f_{\tau'}(x) - f_\tau(x)) \Big).
\end{align*}
By definition, $f_{\tau'}(x) - f_\tau(x) = (\tau' - \tau) d(x)$. Because $\tau' > \tau$ and  $d(x) \ge 0$, the exponential term is upper-bounded by $1$. Therefore, the ratio of the densities is bounded as:
\begin{align} \label{eq:density_to_Z_ratio}
    \frac{p_{\tau'}(x)}{p_\tau(x)} \le \frac{Z_\tau}{Z_{\tau'}}.
\end{align}

We now bound the ratio $Z_\tau / Z_{\tau'}$ using the bounds established in \cref{cor:normalize}. For the numerator, we have
\begin{align*}
    Z_\tau \le \exp(-C_{\tau, \text{low}}) \left(\frac{2\pi}{\mu_\tau}\right)^{n/2} \le \left(\frac{2\pi}{\mu_\tau}\right)^{n/2},
\end{align*}
where the last inequality follows because $C_{\tau, \text{low}} \ge 0$. For the denominator, we have:
\begin{align*}
    Z_{\tau'} \ge  \exp(-C_{\tau', \text{up}}) \left(\frac{2\pi}{L_{\tau'}}\right)^{n/2} \ge \exp(-E_{\text{shift}}) \left(\frac{2\pi}{L_{\tau'}}\right)^{n/2}.
\end{align*}
where we recall from \cref{lem:target_envelope} that $C_{\tau', \text{up}} \le E_{\text{shift}}$. Therefore:
\begin{align*}
    \frac{Z_\tau}{Z_{\tau'}} \le \exp(E_{\text{shift}}) \left( \frac{L_{\tau'}}{\mu_\tau} \right)^{n/2}.
\end{align*}

We first assume $k \ge 1$. We substitute the definitions $L_{\tau'} = \mu_0 + \tau' L_d$ and $\mu_\tau = \mu_0 + \tau \mu_d$, along with the geometric schedule $\tau' = (1+\gamma)\tau$:
\begin{align*}
    \frac{L_{\tau'}}{\mu_\tau} = \frac{\mu_0 + (1+\gamma)\tau L_d}{\mu_0 + \tau \mu_d} = \frac{\mu_0 + \tau L_d}{\mu_0 + \tau \mu_d} + \gamma \frac{\tau L_d}{\mu_0 + \tau \mu_d} \le \kappa_{\max}(1+\gamma).
\end{align*}
where the last inequality follows since $\kappa_{\max} \ge \sup_{\tau} (L_\tau / \mu_\tau)$.
Plugging this back into the partition function ratio gives:
\begin{align*}
    \frac{Z_\tau}{Z_{\tau'}} \le \exp(E_{\text{shift}}) \kappa_{\max}^{n/2} (1+\gamma)^{n/2} \le B (1+\gamma)^{n/2}
\end{align*}

Substituting this back into \cref{eq:density_to_Z_ratio} and multiplying by the second term $r(x) \le K$, we have:
\begin{align*}
    \frac{p_{\tau'}(x)}{\mathcal{L}_\tau(x)} \le K \cdot B \cdot (1+\gamma)^{n/2}.
\end{align*}

When $k = 0$, note that $\tau = 0$ and $\tau' = \mu_0/L_d$, so that 
$$ \frac{L_{\tau'}}{\mu_\tau} = \frac{2 \mu_0}{\mu_0} = 2 \le \kappa_{\max}.$$
Repeating the above argument completes the proof. Note that setting $\gamma = \frac{1}{n}$ yields the bound:
$$ \sup_{x} \frac{p_{\tau'}(x)}{\mathcal{L}_\tau(x)} \le K \cdot B \cdot \left(1 + \frac{1}{n}\right)^{n/2} \le K \cdot B \cdot \sqrt{e}. $$
This justifies the use of $\sqrt{e}KB$ as the bounding constant in the rejection sampling step.
\end{proof}

The previous lemma implies the accept probability in rejection sampling is at most one. We now bound the sampling steps needed to generate an accepted sample.

\begin{lemma}
    In any sampling step, the probability of acceptance is at least $\frac{1}{\sqrt{e}B^2 K}$. Therefore, the probability that the sampling process ever fails over the entire course of the algorithm is at most $\delta$.
\end{lemma}
\begin{proof}
First, the acceptance probability is well-defined. Since $\bar Z_{\tau_{k+1}}\ge Z_{\tau_{k+1}}$, we have $\hat p_{\tau_{k+1}}(x)\le p_{\tau_{k+1}}(x)$ for all $x$. By \cref{lem:geometric_ratio}, $p_{\tau_{k+1}}(x)/\model_{\tau_k}(x)\le \sqrt e K B$, and hence $\frac{1}{\sqrt e K B}\frac{\hat p_{\tau_{k+1}}(x)}{\model_{\tau_k}(x)}\le 1$.

Conditional on acceptance, the accepted sample has density proportional to $\model_{\tau_k}(x)\cdot \frac{1}{\sqrt e K B}\frac{\hat p_{\tau_{k+1}}(x)}{\model_{\tau_k}(x)}$, which is proportional to $\exp(-f_{\tau_{k+1}}(x))$. Thus the accepted sample is distributed exactly according to $p_{\tau_{k+1}}$. It remains to lower-bound the acceptance probability. Averaging over $x\sim \model_{\tau_k}$ gives
\begin{align*}
\Pr[\mbox{Accept}]
&= \int \model_{\tau_k}(x) \frac{1}{\sqrt{e} K B} \frac{\hat{p}_{\tau_{k+1}}(x)}{\model_{\tau_k}(x)} dx
= \frac{1}{\sqrt{e} K B}\int \hat p_{\tau_{k+1}}(x)dx \\
&= \frac{1}{\sqrt{e} K B}\frac{Z_{\tau_{k+1}}}{\bar Z_{\tau_{k+1}}}
\ge \frac{1}{\sqrt{e} K B^2},
\end{align*}
where the last inequality uses $\bar Z_{\tau_{k+1}}/Z_{\tau_{k+1}}\le B$. To see this, note that by the lower bound in \cref{cor:normalize}, $Z_\tau \ge e^{-C_{\tau,\mathrm{up}}}\left(2\pi/\Lambda_\tau\right)^{n/2}$, where $\Lambda_\tau := \mu_0+\tau L_d$. Therefore
$$
\frac{\bar Z_\tau}{Z_\tau}
\le
\frac{\left(2\pi/\mu_\tau\right)^{n/2}}
{e^{-C_{\tau,\mathrm{up}}}\left(2\pi/\Lambda_\tau\right)^{n/2}}
=
e^{C_{\tau,\mathrm{up}}}\left(\frac{\Lambda_\tau}{\mu_\tau}\right)^{n/2}
\le
e^{E_{\rm shift}}\kappa_{\max}^{n/2}
=
B,
$$
where we used $C_{\tau,\mathrm{up}}\le E_{\rm shift}$ and $\Lambda_\tau/\mu_\tau\le \kappa_{\max}$.

Since the rejection process follows a geometric distribution, the probability a sample is not generated in $t_s = 10KB^2 \ln \frac{2Nm}{\delta}$ steps is at most $\exp(-t_s/(\sqrt e K B^2))\le \frac{\delta}{2Nm}$. Summing over $N \cdot m$ total samples across all temperatures yields a failure probability of at most $\delta/2$. Note that in \cref{lem:geometric_ratio}, there is a probability $\delta/(2N)$ per temperature that the density ratio itself is not bounded. Taking the union bound over all such bad events yields a total failure probability of at most $\delta$.
\end{proof}

\subsubsection{Relating Temperature to the Precision of Global Optimization}
We now find the $T$ needed to achieve precision $\mathbb{E}_{p_T}[d(x)]  \le \epsilon$ for a given $\epsilon$, hence casting our sample complexity in more standard terms. Recall we assumed $d(x^*) = 0$.

\begin{lemma}[Precision Scaling] \label{lem:precision}
To achieve an expected objective error $\mathbb{E}_{p_T}[d(x)] \le \epsilon$, it suffices to set the final temperature to $T = O(1/\epsilon)$. This implies an overall sample complexity of $\widetilde{O}(\log 1/\epsilon)$.
\end{lemma}

\begin{proof}
Since $d(x) \le \frac{L_d}{2}\|x - x^*\|^2$,  taking the expectation over the final target distribution $p_T(x)$:
\begin{align*}
    \mathbb{E}_{p_T}[d(x)] \le \frac{L_d}{2} \mathbb{E}_{p_T}[\|x - x^*\|^2] \le L_d \mathbb{E}_{p_T}[\|x - \bar{x}_T\|^2] + L_d \|\bar{x}_T - x^*\|^2,
\end{align*}
where we have used the triangle inequality. The second term expands to:
\begin{align*}
    \|\bar{x}_T - x^*\|^2 = \left\| \frac{\mu_0 x_0 + T \mu_d x^*}{\mu_0 + T \mu_d} - x^* \right\|^2 = \frac{2\mu_0 E_{\text{shift}}}{(\mu_0 + T \mu_d)^2} = O\left( \frac{1}{T^2} \right).
\end{align*}
To bound the term $\mathbb{E}_{p_T}[\|x - \bar{x}_T\|^2]$, we use \cref{lem:target_envelope}. Let $q_T = \mathcal{N}(\bar{x}_T, \mu_T^{-1} I_n)$, where $\mu_T = \mu_0 + T\mu_d$. Since $p_T(x) \le B q_T(x)$ for all $x$, we have
$$
\mathbb{E}_{p_T}[\|x - \bar{x}_T\|^2]
\le B \mathbb{E}_{q_T}[\|x - \bar{x}_T\|^2]
= \frac{nB}{\mu_0 + T\mu_d}
= O\!\left(\frac{1}{T}\right).
$$
For sufficiently large $T$, the $O(1/T)$ dominates, so that the error is $O(\epsilon)$. Recall that the sample complexity is $\widetilde{O}\left(\ln T \cdot \ln \frac{\ln T}{\delta} \right)$, which to the higher order multiplicative term is $\widetilde{O}(\log 1/\epsilon)$.
\end{proof}

\subsection{Comparison to Other Non-convex Optimization Methods}
\label{app:compare_mbo}
Our sample complexity bound decouples the problem parameters such as dimension $n$ from the  precision $\epsilon$, yielding a $\tilde{O}(\log 1/\epsilon)$ sample complexity, with constants depending on problem parameters. We compare this guarantee with extant methods for non-convex optimization.

\textbf{1. Grid Search (Lipschitz Optimization):} Standard spatial discretization algorithms require $O((1/\epsilon)^n)$ function evaluations, so the precision parameter $\epsilon$ gets exponentially modified.

\textbf{2. Local MCMC and Simulated Annealing:} While local Markov chains scale polynomially in dimension on convex bodies, they can take exponential time in non-convex landscapes. The Holley-Stroock perturbation lemma \citep{holley1987logarithmic} implies escaping a local minimum of depth $\Delta E$ at inverse temperature $\tau$ requires mixing time $O(\exp(\tau \Delta E))$, and Hajek's theorem \citep{hajek1988cooling} implies guaranteeing convergence in local MCMC simulated annealing requires a logarithmic cooling schedule $\tau_t = O(\ln t)$. These results mean that in the worst case, classical simulated annealing may require $O(\exp(1/\epsilon))$ steps, so again, the precision parameter is exponentially modified. 

To mitigate these slow mixing times, classical MCMC and simulated annealing often employ variance-inflated distributions, either as heavier-tailed proposals to guarantee ergodicity~\citep{mengersen1996rates} or as ``warm starts'' to bound the density ratio between temperature steps~\citep{lovasz2006simulated}. However, classical methods utilize variance inflation to aid \textit{local} random walks, which can take exponentially many samples for non-convex landscapes. In contrast, \alift adapts the variance-inflation principle to bound the mass coverage of a \textit{generative proxy}. The proxy dynamically absorbs empirical estimation errors independent of the annealing temperature, allowing the algorithm to achieve logarithmic sample complexity.

\textbf{3. Deterministic Optimistic Optimization (DOO):} Among finite-sample methods, the closest comparators are optimistic partitioning methods such as DOO~\citep{bubeck2011x}, SOO~\citep{munos2011optimistic} and their stochastic/noisy variants (e.g. HOO/StoSOO~\citep{pmlr-v28-valko13} and POO~\cite{Munos_bb}). In contrast, classical Lipschitz branch-and-bound methods~\cite{PIYAVSKII197257,Shubert} only convergence in different regimes. When the objective is bounded by a quadratic envelope, spatial-partitioning algorithms~\citep{munos2011optimistic,bubeck2011x} achieve $\widetilde{O}(\log(1/\epsilon))$ sample complexity. These methods hierarchically search the space via optimistic estimates of the objective within grid partitions. \alift achieves this same characteristic rate, but under different trade-offs. Unlike these methods, \alift requires prior estimates of both the upper and lower bounds on the curvature parameters ($\mu_d, L_d$), and its sample complexity scales with the full ambient dimension $n$, whereas space-partitioning bounds scale with the near-optimality dimension, which could be smaller. Furthermore, extensions like SOO~\citep{munos2011optimistic} only require the existence of an upper bounding quadratic without needing to know its parameters, a  setting \alift does not handle. Addressing these limitations in \alift is an exciting direction for future work.

On the other hand, space-partitioning operations are harder to define for discrete, combinatorial, or semantic generative spaces (e.g., sequences generated by LLMs). The \alift algorithm matches the sample complexity of space partitioning methods for quadratic-bounded functions, while generalizing as an algorithmic framework to other generative priors and discrete optimization objectives whenever \cref{ass:learn} holds.


\section{Theoretical Evidence for Coarse Learnability}
\label{sec:justify}
Complementing the sample complexity proof in \cref{app:nonconvex_mbo}, we now present theoretical evidence to support the coarse learnability assumption (\cref{ass:learn}) for simple generative models. First, to demonstrate robustness in the \emph{agnostic} setting, we prove that a structurally misspecified learner (a single Gaussian) naturally satisfies the coverage condition when approximating a multimodal target (\cref{app:miss}). Second, to validate consistency with classical MBO theory, we show that standard MLE in the \emph{realizable} setting of MRAS (i.e., exponential families) satisfies the assumption with high probability (\cref{sec:exp}). Finally, we demonstrate that coarse learnability extends to a simple Kernel Density Estimation (KDE) task, provided the kernel is deliberately ``over-smoothed'' to act as a robust coverage envelope (\cref{app:kde}). 

\subsection{Gaussian Mixtures and Robustness to Mis-specification}
\label{app:miss}
\cref{ass:learn} particularly relevant in the agnostic or misspecified setting, where the learner cannot model the complex shape of the target but can learn to cover its support. 
Indeed, when a lower-capacity model (like a restricted neural network or a single Gaussian) approximates a complex, multimodal target, the MLE objective forces the learner to ``spread out'' and envelope the target's support rather than collapsing to a single mode~\citep{minka2005divergence}. 
In our setting, the generative model learns to act as this relaxed envelope. 

We illustrate this with a Gaussian mixture model target and a single Gaussian learner. Let the target $p(x)$ be a mixture of $k$ unit-variance Gaussians:
\[ p(x) = \frac{1}{k} \sum_{i=1}^k \mathcal{N}(\mu_i, 1). \]
Let the learner $\mathcal{L}$ be restricted to the family of single Gaussians $\mathcal{N}(\theta, \sigma^2)$. This is a mis-specified setting where $d_{\mathrm{TV}}(p, \mathcal{L})$ cannot be made arbitrarily small. However, we show that MLE naturally satisfies \cref{ass:learn}.
\begin{theorem} \label{thm:exp10}
Let the target $p(x) = \frac{1}{k} \sum_{i=1}^k \mathcal{N}(\mu_i, 1)$ be a mixture of $k$ unit-variance Gaussians with means $\mu_i \in [-\frac{\Delta}{2}, \frac{\Delta}{2}]$, where $\Delta \ge 2$. Assume the modes are sufficiently separated such that the population variance satisfies $\sigma^{*2} \ge 2$. Let $\mathcal{L}_{\mathrm{MLE}}$ be the single Gaussian model $\mathcal{N}(\hat{\mu}, \hat{\sigma}^2)$ learned via MLE on $m$ i.i.d.\ samples from $p$.

If $m = \Omega(\Delta^4 \ln(1/\delta))$, then with probability at least $1 - \delta$, the learned model satisfies the global coverage condition:
\[ \sup_{x \in \mathbb{R}} \frac{p(x)}{\mathcal{L}_{\mathrm{MLE}}(x)} = O(\Delta \cdot e^{k}) := K. \]
Thus, for a fixed number of modes $k$, the model family satisfies \cref{ass:learn} with $m$  scaling polynomially in $K = O(\Delta)$ and $\ln (1/\delta)$.
\end{theorem}

\begin{proof}
Let $\mu^*$ and $\sigma^{*2}$ denote the true mean and variance of the target mixture $p$.

The mean of the mixture is $\mu^* = \frac{1}{k}\sum \mu_i$. The variance of the mixture is given by the law of total variance:
\[ \sigma^{*2} = 1 + \frac{1}{k} \sum_{j=1}^k (\mu_j - \mu^*)^2. \]
From this identity, we observe that the squared deviation of any single component mean is bounded by the total variance scaled by $k$:
\begin{equation}
    \label{eq:var_bound}
    (\mu_i - \mu^*)^2 \le \sum_{j=1}^k (\mu_j - \mu^*)^2 = k(\sigma^{*2} - 1).
\end{equation}
By the premise, $\sigma^{*2} \ge 2$. Since the means are bounded in width $\Delta$, we also have $\sigma^{*2} \le 1 + \Delta^2$.

The MLE parameters are the sample mean $\hat{\mu}$ and sample variance $\hat{\sigma}^2$. The target distribution $p$ is a mixture of Gaussians with bounded parameter support. Since the component means are bounded by $\Delta/2$ and the component variances are fixed at 1, the random variable $X \sim p$ is sub-Gaussian with norm $\|X\|_{\psi_2} = O(\Delta)$. First, by Hoeffding's inequality \citep{vershynin_high-dimensional_2018}, for any $t > 0$ (where $c$ is a constant). 
    \[ \Pr\left(|\hat{\mu} - \mu^*| \ge t\right) \le 2 \exp\left(-c m \frac{t^2}{\Delta^{2}}\right). \]
Next, the variable $X^2$ is sub-exponential with $\|X^2\|_{\psi_1} = O( \Delta^{2})$, so by Bernstein's inequality ~\citep{vershynin_high-dimensional_2018}:
    \[ \Pr\left(|\hat{\sigma}^2 - \sigma^{*2}| \ge t\right) \le 2 \exp\left(-c m \min\left(\frac{t^2}{\Delta^{4}}, \frac{t}{\Delta^{2}}\right)\right). \]
We choose a sampling error tolerance $\eta$ as a sufficiently small constant. To ensure $|\hat{\mu} - \mu^*| \le \eta$ and $|\hat{\sigma}^2 - \sigma^{*2}| \le \eta$ with probability $1-\delta$,  we need $m = \Omega(\Delta^4 \ln(1/\delta))$.
Conditioned on this high-probability event, and using $\sigma^{*2} \ge 2$, we have $\hat{\sigma}^2 > 1$. 

The coverage ratio is $r(x) = \frac{p(x)}{\mathcal{L}(x)} = \frac{1}{k} \sum_{i=1}^k \frac{\mathcal{N}(x; \mu_i, 1)}{\mathcal{N}(x; \hat{\mu}, \hat{\sigma}^2)}$. It suffices to bound an arbitrary term $T_i(x) = \frac{\mathcal{N}(x; \mu_i, 1)}{\mathcal{N}(x; \hat{\mu}, \hat{\sigma}^2)}$:
\[ T_i(x) = \hat{\sigma} \exp\left( -\frac{(x-\mu_i)^2}{2} + \frac{(x-\hat{\mu})^2}{2\hat{\sigma}^2} \right). \]
Let $E_i(x)$ be the exponent. Differentiating with respect to $x$:
\[ E_i'(x) = -(x-\mu_i) + \frac{(x-\hat{\mu})}{\hat{\sigma}^2}. \]
Setting $E_i'(x) = 0$ yields the unique maximum at:
\[ x^* = \frac{\hat{\sigma}^2 \mu_i - \hat{\mu}}{\hat{\sigma}^2 - 1}. \]
Substituting this back into $E_i(x)$, the exponent at the maximum simplifies to:
\[ E_i(x^*) = \frac{(\mu_i - \hat{\mu})^2}{2(\hat{\sigma}^2 - 1)}. \]

We now apply the finite sample bounds derived above.
First, consider the numerator $(\mu_i - \hat{\mu})^2$. Since $|\hat{\mu} - \mu^*| \le \eta$, we have
\[ (\mu_i - \hat{\mu})^2 \le (\mu_i - \mu^*)^2 + 2\eta|\mu_i - \mu^*| + \eta^2. \]
Using the population variance bound from Eq.~\eqref{eq:var_bound}, let $V = \sigma^{*2} - 1$. Then $(\mu_i - \mu^*)^2 \le kV$, which implies $|\mu_i - \mu^*| \le \sqrt{kV}$. Thus:
\[ (\mu_i - \hat{\mu})^2 \le kV + 2\eta\sqrt{kV} + \eta^2. \]

Next, consider the denominator $2(\hat{\sigma}^2 - 1)$. Using $|\hat{\sigma}^2 - \sigma^{*2}| \le \eta$:
\[ \hat{\sigma}^2 - 1 \ge \sigma^{*2} - 1 - \eta = V - \eta. \]
Since we assumed $\sigma^{*2} \ge 2$, we have $V \ge 1$. Since $\eta \le 0.5$, we have $V - \eta \ge V(1 - \eta) \ge 0.5 V$.
For small enough constant $\eta$, the exponent $E_i(x^*)$ is bounded by:
\[ E_i(x^*) \le \frac{kV + 2\eta\sqrt{kV} + \eta^2}{2(V - \eta)} = \frac{k}{2} \left( \frac{1 + \frac{2\eta}{\sqrt{kV}} + \frac{\eta^2}{kV}}{1 - \frac{\eta}{V}} \right) \le k. \]
Consequently, the maximum value of the component ratio is:
\[ \sup_x T_i(x) = \hat{\sigma} \exp(E_i(x^*)) \le \hat{\sigma} \cdot e^{k}. \]
 Since $\hat{\sigma}^2 \le \sigma^{*2} + \eta \le 2 + \Delta^2$, we have $\hat{\sigma} \le O(\Delta)$.

Summing over the $k$ components (each with weight $1/k$):
\[ \sup_x \frac{p(x)}{\mathcal{L}(x)} \le \frac{1}{k} \sum_{i=1}^k \sup_x T_i(x) \le \frac{1}{k} \sum_{i=1}^k O(\Delta \cdot e^{k}) = O(\Delta \cdot e^{k}). \]
We can now choose any $m \ge \mathrm{poly}\left( \ln \frac{1}{\delta}, \Delta, e^{k} \right)$ to satisfy \cref{ass:learn}. Note that the coverage condition holds globally.
\end{proof}

\subsection{Exponential Families and Realizable Setting}
\label{sec:exp}
We now present theoretical justification for coarse learnability for simple distributions. We consider the class of exponential families with a single parameter and show that under  benign assumptions on the family, the MLE estimation problem is coarsely learnable. This result extends to some multi-parameter settings, and we present a sketch at the end omitting the details. Note that single-parameter exponential families capture most common distributions such as Gaussian, Exponential, Poisson, Binomial, etc. This result shows that coarse learnability is a natural property of standard statistical estimation in well-behaved settings. 

It is important to contextualize the result below within the model reference adaptive search (MRAS) literature. As discussed in \cref{sec:mras_connection}, standard MRAS utilizes natural exponential families for the parametric model~\citep{hu2007model}. While true target distributions are generally complex, analyzing the idealized ``realizable'' setting, where the target falls within the model family, serves as a theoretical baseline. The analysis below confirms that in this realizable setting, the finite-sample MLE inherently satisfies our coarse learnability assumption. This indicates that our statistical condition is well-grounded, holding for standard parametric families in an idealized realizable setting. Crucially, however, as discussed in \cref{app:miss}, our framework extends guarantees beyond this idealized case to the more general agnostic setting where the target is complex and the model is misspecified, by utilizing the Metropolis--Hastings correction.

\paragraph{Single-Parameter Setting.} Let $p(x \mid \eta) = h(x) \cdot e^{\eta x - A(\eta)}$ be a single-parameter exponential family and let $X$ denote a random variable following this distribution. Let $\mu(\eta) = \E_{X \sim p(x \mid \eta)}[X] = A'(\eta)$. We suppose the exponential family  satisfies the following simple assumption. 

\begin{assume}[Regularity] \label{asm:regularity_thm}
We assume the density $p(x \mid \eta)$ is sub-exponential.\footnote{A sub-exponential random variable satisfies $\Pr[|X| \ge K] \le 2 e^{-C \cdot K}$ for all $K \ge 0$ and some constant $C > 0$.} Further, the true parameter $\eta_0$ lies in a constant-sized closed interval $\Omega_0$ such that for all $\eta \in \Omega_0$, we have that $A(\eta)$ is analytic and $0 < v_{\min} \le A''(\eta) \le v_{\max} < \infty$ for some positive constants $v_{\min}, v_{\max}$.
\end{assume}

The above assumption holds for common distributions such as Gaussian, Bernoulli, Exponential, and Poisson under mild conditions on their parameters. These densities are clearly sub-exponential. Furthermore, we have the following:

\begin{description}
\item[Gaussian (with known variance).] We have $p(x \mid \mu) = \frac{1}{\sqrt{2\pi\sigma^2}} e^{-(x-\mu)^2/(2\sigma^2)}$. Therefore, $\eta = \mu/\sigma^2$ and $A(\eta) = \eta^2 \sigma^2 / 2$, so that $A''(\eta) = \sigma^2$, which is a constant.
\item[Bernoulli.] We have $p(x \mid \mu) = \mu^x (1-\mu)^{1-x}$ for $x \in \{0,1\}$. We have $\eta = \ln(\mu/(1-\mu))$, and $A(\eta) = \ln(1+e^{\eta}) = -\ln(1-\mu)$.  We have $A''(\eta) = e^\eta / (1+e^\eta)^2 = \mu(1-\mu)$, which satisfies the above assumption when $\mu \in [\delta, 1-\delta]$ for constant $\delta > 0$.
\item[Exponential.] We have $p(x \mid \lambda) = \lambda e^{-\lambda x}$ for $x \ge 0$. Further, we have $\eta = -\lambda$ and $A(\eta) = - \ln(-\eta)$, so that $A''(\eta) = 1/\eta^2$, which satisfies the above assumption when $\lambda > 0$ is a constant.
\item[Poisson.] We have $p(x \mid \lambda) = e^{-\lambda} \lambda^x / x!$ for $x=0,1,2,\ldots$. We have $\eta = \ln \lambda$, $A(\eta) = e^{\eta}$, and $A''(\eta) = e^{\eta} = \lambda$, which satisfies the assumption when $\lambda > 0$ is a constant.
\end{description}

The derivation of coarse learnability under \cref{asm:regularity_thm} uses standard properties of sub-exponential distributions; see, e.g.,~\citep{vershynin_high-dimensional_2018} for details. Since we assumed the exponential family is sub-exponential, this means there exist positive constants $C_1, C_2 > 0$ such that for $X \sim p(x \mid \eta_0)$:
\begin{equation} 
\label{eq:sufficient}
\pr{\abs{X - \mu(\eta_0)} > \theta} \le C_1 e^{-C_2 \cdot \theta}. 
\end{equation}
This also means~\citep{vershynin_high-dimensional_2018} that there are constants  $\nu^2, \alpha$ such that for random variable $X$ following $p(x \mid  \eta)$, we have:
\begin{equation*}
\E[e^{\lambda X}] \le e^{\lambda^2 \nu^2/2} \qquad \forall |\lambda| \le \frac{1}{\alpha}.
\end{equation*}
Let $Z = X - \mu(\eta)$, so that $\E[Z] = 0$. Clearly, $Z$ is also sub-exponential, and satisfies the above equation for some $\nu^2, \alpha$. Suppose we draw $m$ samples from $p(x \mid  \eta)$ and let $\hat{X}$ be their average:
\begin{align*}
\pr{\hat{X} - \mu(\eta) > \epsilon}  = & \pr{\sum_{i=1}^m Z_i > m \cdot \epsilon} 
 = \pr{\prod_{i=1}^m e^{\lambda Z_i} > e^{m \cdot \lambda \cdot \epsilon}}  \\
 \le & \frac{\left(\E[e^{\lambda Z}]\right)^m}{e^{m \cdot \lambda \cdot \epsilon}} 
 \le  e^{m \left(\lambda^2  \cdot \nu^2/2 - \lambda \cdot \epsilon\right).} 
\end{align*}
Choosing $\lambda = \frac{\epsilon}{\nu^2} < \frac{1}{\alpha}$ (assuming $\epsilon = o(1)$), we have
$$ \pr{\abs{\hat{X} - \mu(\eta)} > \epsilon}  \le 2 e^{- \frac{m \cdot \epsilon^2}{2 \nu^2} }.$$
 Choosing $\epsilon  = m^{-1/3}$, this implies:
\begin{equation}
\label{eq:conc}
 \pr{\abs{\hat{X} - \mu(\eta)} > m^{-1/3}} \le 2 \exp\left(-\frac{m^{1/3}}{2 \nu^2}\right). 
\end{equation}

We are now ready to show the following theorem, which we show implies \cref{ass:learn}.

\begin{theorem}
\label{thm:exp1}
Suppose the exponential family satisfies \cref{asm:regularity_thm}. Let $\hat{\eta}_m$ be the MLE of $\eta_0$ based on $m$ samples, satisfying $A'(\hat{\eta}_m) = \bar{X}_m$, where $\bar{X}_m$ is the average of $m$ samples. Let $\L(x) \equiv p(x \mid \hat{\eta}_m)$.
For any small constant $\gamma > 0$, define 
$$W_m = \left\{x : \abs{\ln p(x \mid \eta_0) - \ln p(x \mid \hat{\eta}_m)} \le m^{-1/3 + \gamma} \right\}.$$
Then, for any $m \ge \mathrm{poly}\left(\frac{v_{\max}}{v_{\min}}, \frac{1}{\nu}\right)$ and a sufficiently small constant $\gamma > 0$, with probability $1 - 2 e^{- m^{\gamma}}$, the parameter $\hat{\eta}_m$ satisfies $ \Pr[X \notin W_m] \le e^{-m^{\gamma}}.$
\end{theorem}
\begin{proof}
Let $\Delta\eta = \hat{\eta}_m - \eta_0$. The log-density ratio is
$$\mathcal{R}(x) = \ln p(x \mid \eta_0) - \ln p(x \mid \hat{\eta}_m) = (\eta_0 - \hat{\eta}_m)\cdot x + (A(\hat{\eta}_m) - A(\eta_0)).$$
Using a Taylor expansion for $A(\hat{\eta}_m) = A(\eta_0 + \Delta\eta)$ around $\eta_0$ up to the second term, for some $\xi$ between $\eta_0$ and $\hat{\eta}_m$, we have 
$$A(\hat{\eta}_m) - A(\eta_0)   = A'(\eta_0)\Delta\eta + \frac{1}{2}A''(\xi)(\Delta\eta)^2.$$
Substituting this into $\mathcal{R}(x)$:
$$
\mathcal{R}(x) = \Delta\eta (A'(\eta_0) - x) + \frac{1}{2}A''(\xi)(\Delta\eta)^2. 
$$

Let $\mu(\eta_0) = A'(\eta_0)$ be the true mean. By \cref{eq:conc}, the event $\mathcal{E}_{\mathrm{param}} := \{\abs{\bar{X}_m - \mu(\eta_0)} \le m^{-1/3}\}$ occurs with probability at least $1 - \exp(-m^{1/3}/\nu^2)$.
The MLE satisfies $A'(\hat{\eta}_m) = \bar{X}_m = \mu(\hat{\eta}_m)$.
On $\mathcal{E}_{\mathrm{param}}$, we therefore have $\abs{A'(\hat{\eta}_m) - A'(\eta_0)} \le m^{-1/3}$.
By the Mean Value Theorem, $A'(\hat{\eta}_m) - A'(\eta_0) = A''(\eta^*)(\hat{\eta}_m - \eta_0)$ for some $\eta^*$ between $\eta_0$ and $\hat{\eta}_m$.
By \cref{asm:regularity_thm}, $A''(\eta^*) \ge v_{\min} > 0$.
Thus, on $\mathcal{E}_{\mathrm{param}}$, we have
$$\abs{\Delta\eta} = \abs{\hat{\eta}_m - \eta_0} \le \frac{1}{v_{\min}} \abs{A'(\hat{\eta}_m) - A'(\eta_0)} \le \frac{1}{v_{\min}} m^{-1/3}.$$
This implies $(\Delta\eta)^2 \le \frac{m^{-2/3}}{v_{\min}^2}$. Also, on $\mathcal{E}_{\mathrm{param}}$,  $A''(\xi) \le v_{\max}$ by \cref{asm:regularity_thm}. This implies the first term $\Delta\eta (A'(\eta_0) - x)$ dominates in the equation for $\mathcal{R}(x)$ when $m \ge \mathrm{poly}\left(\frac{v_{\max}}{v_{\min}}\right)$. Thus, for $\abs{\mathcal{R}(x)} \le m^{-1/3 + \gamma}$, we  require:
$$\abs{\Delta\eta (A'(\eta_0) - x)} \le \frac{1}{2} m^{-1/3 + \gamma} .$$
Since $\abs{\Delta\eta} \le \frac{m^{-1/3}}{v_{\min}}$, this requires 
$\abs{A'(\eta_0) - x} \le \frac{1}{2} v_{\min} m^{\gamma} := K_m$. 
The set $W$ is then defined by $x$ such that $\abs{x - \mu(\eta_0)} \le K_m$. By \cref{eq:sufficient}, we have $\Pr[x \notin W] = O\left(e^{- O(K_m)}\right)$. Choosing $\gamma > 0$ an appropriately small constant, this completes the proof.  
\end{proof}

Note that the failure probabilities in the above theorem are monotonically decreasing in $m$. Therefore, if we are given $K > 1, \epsilon > 0, \delta > 0$ as in \cref{ass:learn}, then we can simply choose a large enough $m \ge \mathrm{poly}\left(\frac{v_{\max}}{v_{\min}}, \frac{1}{\nu}, \ln \frac{1}{\epsilon \delta}\right)$ that ensures $e^{-m^{\beta}} \le \min\left( \delta, \epsilon\right)$. Further, note that $m^{-1/3 + \gamma} < \ln K$. This ensures that with probability $1-\delta$, we have $\Pr \left[p(x | \eta_0) \ge K \cdot p(x | \hat{\eta}_m) \right] \le \epsilon$, hence satisfying \cref{ass:learn}.

\paragraph{Multi-Parameter Setting.} Let $p(\mathbf{x} \mid \boldsymbol{\eta}) = h(\mathbf{x}) \exp(\boldsymbol{\eta}^T \mathbf{x} - A(\boldsymbol{\eta}))$ be a $k$-parameter exponential family, where $\boldsymbol{\eta} \in \Re^k$ and $\mathbf{x} \in \Re^k$, and $k$ being a constant. This for instance, captures multi-dimensional Gaussian distributions with a known covariance matrix, or multinomial distributions. Assuming the resulting density is sub-exponential in the $\ell_2$-norm, it can be shown that such a distribution satisfies \cref{eq:sufficient} and \cref{eq:conc} with the absolute value replaced by the $\ell_2$-norm, and the constants depending on $k$. In particular, the $\ell_2$-norm of the deviation from the mean is a sub-exponential random variable. By re-working the same proof as that of \cref{thm:exp1}, this implies an analog of \cref{thm:exp1} to the multi-dimensional setting, under suitable regularity assumptions on $A(\boldsymbol{\eta})$. The details are easy to fill in, and omitted.

\subsection{Over-smoothed Non-Parametric Estimation}
\label{app:kde}

The previous sections demonstrated that coarse learnability arises naturally in parametric settings due to the mass-covering properties of Maximum Likelihood Estimation. We now show that this principle extends to a simple non-parametric Kernel Density Estimation (KDE) task, provided the estimator is deliberately ``over-smoothed.''

In classical density estimation, the bandwidth $h$ of a kernel is typically shrunk as the sample size $m \to \infty$ to recover the exact target density. However, in our optimization framework, a shrinking bandwidth creates light tails that exponentially under-cover the target's support, violating \cref{ass:learn}. Conversely, if we fix a kernel whose bandwidth dominates the target's tail decay, the KDE forms a robust coverage envelope. We formalize this in $d$-dimensions.

\begin{theorem} \label{thm:kde}
Let the target distribution $p(x) = \mathcal{N}(0, \sigma^2 I_d)$ be a $d$-dimensional isotropic Gaussian. Let $\mathcal{L}_{\text{KDE}}(x) = \frac{1}{m} \sum_{i=1}^m K_h(x - X_i)$ be the density learned via KDE using a Gaussian kernel $K_h(z) = (2\pi h^2)^{-d/2} \exp(-\|z\|^2 / 2h^2)$ on $m$ i.i.d. samples $X_1, \ldots, X_m$ from $p$.

Assume we set the bandwidth to match\footnote{The proof easily extends to $h = c \cdot \sigma$ for constant $c \ge 1$.} the target's tail decay, $h = \sigma$. There exists a universal constant $C > 0$ such that if the sample size satisfies $m \ge C(d + \ln(1/\delta))$, then with probability at least $1 - \delta$ over the samples, the learned model globally satisfies the coarse learnability condition:
\begin{align*}
    \sup_{x \in \mathbb{R}^d} \frac{p(x)}{\mathcal{L}_{\text{KDE}}(x)} \le 8 e^d := K.
\end{align*}
This implies $m = O\left(\ln{\frac{K}{\delta}}\right)$, showing \cref{ass:learn}.
\end{theorem}

\begin{proof}
We first show that for any query point $x \in \mathbb{R}^d$, a constant fraction of the empirical samples simultaneously resides near the origin and directionally points toward $x$.

Let $\mathcal{B} = \{y \in \mathbb{R}^d : \|y\|^2 \le 2\sigma^2 d\}$ be the closed ball of radius $\sigma\sqrt{2d}$ centered at the origin. For $X \sim \mathcal{N}(0, \sigma^2 I_d)$, the expected squared norm is $\mathbb{E}[\|X\|^2] = \sigma^2 d$. By Markov's inequality, the probability that a sample falls outside this ball is bounded by:
\begin{align*}
    \Pr(\|X\|^2 \ge 2\sigma^2 d) \le \frac{\mathbb{E}[\|X\|^2]}{2\sigma^2 d} = \frac{1}{2}
\end{align*}
Thus, the true probability mass of the ball is $\Pr(X \in \mathcal{B}) \ge 1/2$.

For any query point $x \in \mathbb{R}^d$, define the halfspace pointing toward $x$ as $H_x = \{y \in \mathbb{R}^d : \langle x, y \rangle \ge 0\}$. Because the Gaussian distribution is spherically symmetric and centered at the origin, exactly half of the mass of $\mathcal{B}$ lies in $H_x$. Therefore, the true probability that a random sample falls into their intersection is:
\begin{align*}
    p_x := \Pr(X \in \mathcal{B} \cap H_x) = \frac{1}{2} \Pr(X \in \mathcal{B}) \ge \frac{1}{4}
\end{align*}

We require the empirical fraction of samples in $\mathcal{B} \cap H_x$ to tightly concentrate around its true expectation uniformly over all possible directions $x \in \mathbb{R}^d$. The set of such halfspaces $\{H_x : x \in \mathbb{R}^d\}$ has VC dimension of $d$. Intersecting this class with a fixed set $\mathcal{B}$ does not increase the VC dimension. By uniform convergence, there exists a universal constant $c > 0$ such that with probability at least $1 - \delta$:
\begin{align*}
    \sup_{x \in \mathbb{R}^d} \left| \frac{1}{m} \sum_{i=1}^m \mathbf{1}_{\{X_i \in \mathcal{B} \cap H_x\}} - p_x \right| \le c \sqrt{\frac{d + \ln(1/\delta)}{m}}
\end{align*}
We choose the sample size $m$ sufficiently large such that this uniform error is bounded by $1/8$. Specifically, this requires $m \ge C(d + \ln(1/\delta))$ for $C = 64c^2$.

Under this condition, with probability $1 - \delta$, the empirical fraction of samples falling in $\mathcal{B} \cap H_x$ is at least $p_x - 1/8 \ge 1/4 - 1/8 = 1/8$ for all $x$ simultaneously. Let $S_x = \{X_i : X_i \in \mathcal{B} \cap H_x\}$ be this subset of samples for a given $x$. We are guaranteed that $|S_x| \ge m/8$ for every $x \in \mathbb{R}^d$.

Conditioned on this uniform convergence event, we evaluate the ratio $r(x) = p(x) / \mathcal{L}_{\text{KDE}}(x)$ for an arbitrary $x \in \mathbb{R}^d$. We lower bound the KDE sum by dropping all kernel components except those corresponding to the samples in $S_x$:
\begin{align*}
    \mathcal{L}_{\text{KDE}}(x) = \frac{1}{m} \sum_{i=1}^m K_h(x - X_i) \ge \frac{1}{m} \sum_{X_i \in S_x} \frac{1}{(2\pi \sigma^2)^{d/2}} \exp\left( - \frac{\|x - X_i\|^2}{2\sigma^2} \right)
\end{align*}
For every $X_i \in S_x$, we know two  facts by definition of the set: $\|X_i\|^2 \le 2\sigma^2 d$ and $\langle x, X_i \rangle \ge 0$. Expanding the squared distance yields:
\begin{align*}
    \|x - X_i\|^2 = \|x\|^2 - 2\langle x, X_i \rangle + \|X_i\|^2 \le \|x\|^2 + 2\sigma^2 d
\end{align*}
Substituting this upper bound on the distance into each kernel exponent in the sum gives:
\begin{align*}
    \mathcal{L}_{\text{KDE}}(x) &\ge \frac{1}{m} \sum_{X_i \in S_x} \frac{1}{(2\pi \sigma^2)^{d/2}} \exp\left( - \frac{\|x\|^2}{2\sigma^2} \right) \exp(-d) \\
    &= \frac{|S_x|}{m} \frac{1}{(2\pi \sigma^2)^{d/2}} \exp\left( - \frac{\|x\|^2}{2\sigma^2} \right) \exp(-d)
\end{align*}
Because $|S_x| \ge m/8$ for all $x$, we have
\begin{align*}
    \mathcal{L}_{\text{KDE}}(x) \ge \frac{1}{8} \exp(-d) \cdot p(x) \qquad \Rightarrow \qquad
    \sup_{x \in \mathbb{R}^d} \frac{p(x)}{\mathcal{L}_{\text{KDE}}(x)} \le 8 e^d := K
\end{align*}
This completes the proof.
\end{proof}

Our proof easily extends to the setting where $h = c \cdot \sigma$ for constant $c \ge 1$ and yields $K = O( (ce)^d)$. Further, the same proof idea generalizes to centrally symmetric sub-Gaussian distributions with similar bounds.

\section{Empirical Illustration of \alift versus \topift}
\label{sec:empirical_10d}

Although our contributions are primarily theoretical, we present a focused empirical demonstration to validate the necessity of an optimistic, variance-inflated proposal distribution combined with sample correction. We compare \alift against \topift, a baseline that relies on empirical variance without distribution correction and consequently suffers from premature convergence to local optima. While specialized heuristics exist to navigate specific non-convex landscapes, our objective here is not to achieve state-of-the-art empirical performance. Rather, our goal is to cleanly isolate and verify how the core theoretical insights of our framework hold up under strict sample budgets.

\begin{figure}[htbp]
    \centering
      \begin{minipage}[b]{0.66\textwidth}
    \centering
    \includegraphics[width=\linewidth]{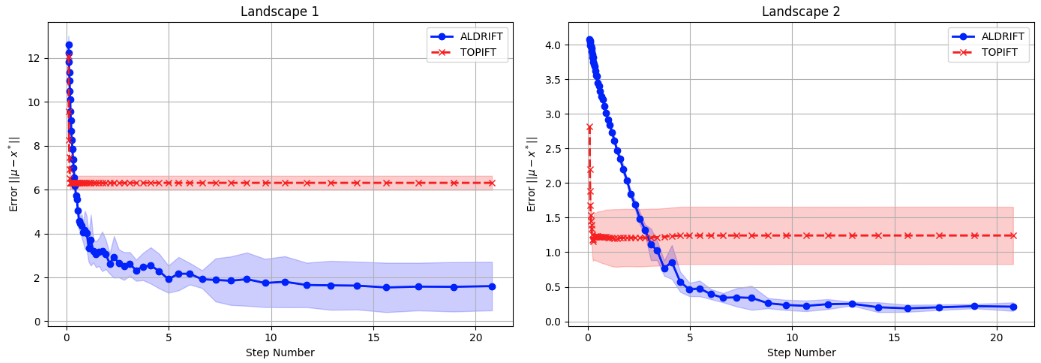} 
    \end{minipage}
          \begin{minipage}[b]{0.33\textwidth}
    \centering
    \includegraphics[width=\linewidth]{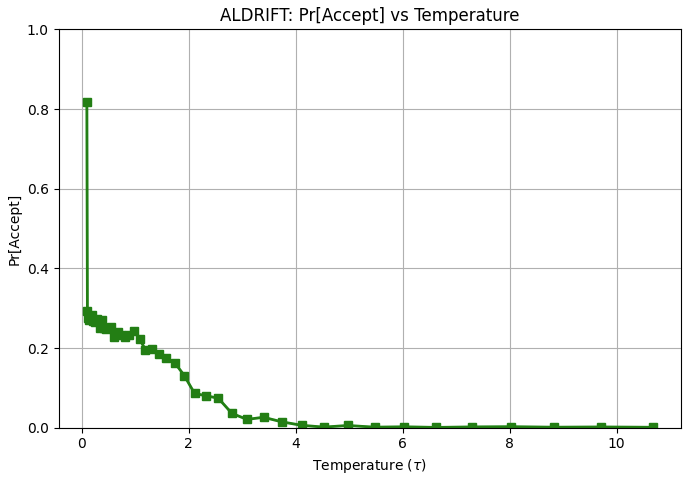}
    \end{minipage}
    \caption{Optimization error trajectories ($||\mu - x^*||$) versus iteration for \alift and \topift. 
    \textbf{(Left)} Parameters: $\mu_d = 0.5, A = 2.0, \omega = 3.0, x_0 = \mathbf{6}, \mu_0 = 0.1$. 
    \textbf{(Center)} Parameters: $\mu_d = 1.0, A = 0.5, \omega = 4.0, x_0 = \mathbf{1.35}, \mu_0 = 2.0$. 
    \textbf{(Right)} $\Pr[\mbox{Accept}]$ of a sample  versus the temperature $\tau$ in \alift.}
    \label{fig:10d_optimization}
\end{figure}

\noindent {\bf Objective Landscape.}
We study the non-convex optimization setting from \cref{app:nonconvex_mbo} and define a $n=10$ dimensional objective $d(\cdot)$ characterized by a global quadratic envelope and high-frequency local minima. For $x \in \mathbb{R}^{10}$, let
$ d(x) = \frac{\mu_d}{2} \|x\|^2 + A \sum_{i=1}^{10} (1-\cos(\omega x_i)) $.
The global minimum is located at $x^* = \mathbf{0}$. We initialize the prior $\mathcal{L}_0 = \mathcal{N}(x_0, \frac{2}{\mu_{0}}I_{n})$ with $x_0$ far enough that the optimizer must traverse multiple local optima to succeed.  Following the setting in \cref{app:nonconvex_mbo} assume $\mu_d$ is known to \alift as a lower bound on curvature at the optimum solution.

\noindent {\bf Algorithms.}
We evaluate both \alift and \topift using a shared sample budget.
\begin{itemize}
    \item \textbf{\alift (Variance Inflated):} We employ the geometric annealing schedule (as described in \cref{app:nonconvex_mbo}), updating the temperature as $\tau_{k+1} = (1 + 1/n)\tau_k$ from an initial $\tau_0 = 0.1$ to $T = 20$. At each step, we use Metropolis-Hastings to draw $m = 200$ samples, with a chain length of $M = 50$. Crucially, the model is updated using the inflated variance $\mathcal{L}_{\tau}(x) = \mathcal{N}(\hat{\omega}_{\tau}, \frac{2}{\mu_{\tau}}I_{n})$, where $\mu_\tau = \mu_0 + \tau \mu_d$ and $\hat{\omega}_{\tau}$ is the sample mean.
    \item \textbf{\topift (Empirical Variance):} The heuristic samples $m \cdot M$ times directly from the current Gaussian model, scores them based on $d(x)$, selects the top $m$ samples, and fits the next Gaussian model using their empirical mean and empirical covariance.
\end{itemize}


\noindent {\bf Results.}
Figure \ref{fig:10d_optimization} (left and center figures) plots the error of the sample mean of the model $\model_{\tau}$ as a function of iteration number for two different settings of the parameters, averaged over $10$ runs for each algorithm. \topift falls into a sub-optimal local minimum because the elite samples are concentrated in a single valley and the empirical variance shrinks to near-zero. Conversely, \alift successfully converges close to the global optimum by using an inflated proposal distribution and correcting for it every iteration, empirically validating \cref{thm:funnel_mbo}. 

Observe that \alift converges \emph{close} to the global optimum, but does not reach it exactly. This occurs because we operate in a highly constrained sampling regime where $M$ is much smaller than the theoretical requirement for convergence. Consequently, as the temperature increases and the target distribution sharpens, the probability of an accepting M-H transition approaches zero (see \cref{fig:10d_optimization}, Right). (A similar stalling effect is observed in our LLM experiments in \cref{sec:empirical}.) Thus, the primary practical advantage of \alift lies in its ability to rapidly isolate the global basin of attraction using a limited sample budget. Once there, standard local optimization methods can be applied to precisely locate the exact optimum. This contrasts with \topift, which frequently becomes trapped in suboptimal local minima under comparable sample constraints. Conversely, in the asymptotic regime where sample size is unconstrained, naive exploration strategies such as grid search or \topift become competitive by brute-forcing the objective landscape, analogous to the asymptotic convergence guarantees of the Cross-Entropy method and MRAS \cite{rubinstein1999cross,hu2007model}.


\section{Motivating Application: Algorithm-LLM Interaction at Inference-time}
\label{app:extended_motivation}

\begin{figure}[htbp]
\begin{center}
        \includegraphics[width = 0.5\linewidth]{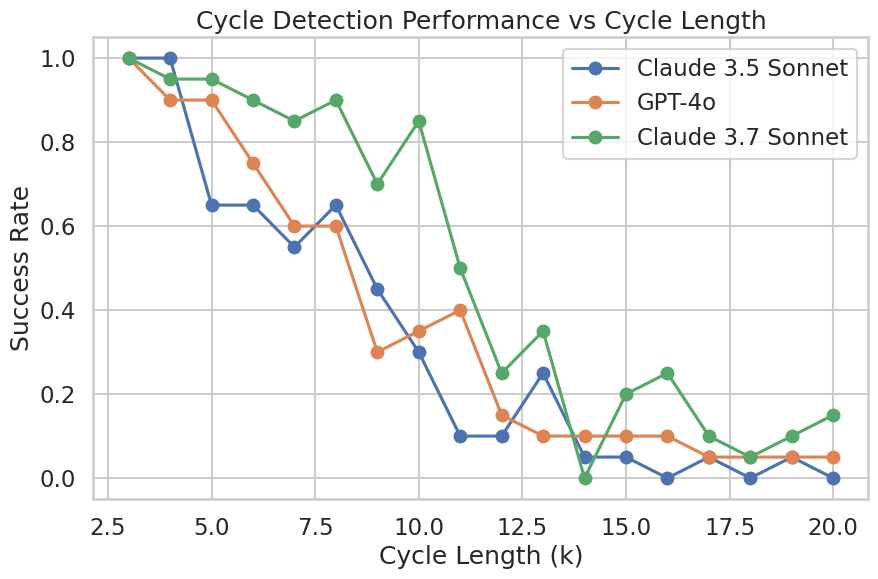}
\end{center}
\caption{\label{fig:cycle} Success rate of the models in finding a length $k$-cycle in a graph as a function of $k$. For each $k$, the graph is constructed by starting with a cycle of length $k$ and randomly adding $\lceil k/2 \rceil$ other edges. For each cycle length $k$, we generate 50 random instances and compute the success rate, i.e., the fraction of times the model returns a length-$k$ cycle. See Appendix~\ref{app:prompts} for the exact prompt used.}
\end{figure}

One of our motivations for studying MBO with expressive, black-box generative priors comes from algorithm-LLM interaction, specifically via inference-time alignment. Increasingly, real-world combinatorial optimization requires balancing strict global constraints (e.g., graph connectivity, efficiency) with informal, context-dependent local specifications (e.g., scenic preferences, stylistic coherence). While classical algorithms efficiently solve for global feasibility, they are brittle when handling qualitative or ambiguous requirements~\citep{wang2024languagemodelssolvegraph}. Similarly, while modern LLMs excel at interpreting open-ended requirements~\cite{radford2018improving, brown2020languagemodelsfewshotlearners}, they consistently struggle to enforce global combinatorial properties. We illustrate this limitation empirically using a cycle detection task. As shown in Figure~\ref{fig:cycle}, when asked to find a length-$k$ cycle in a graph containing a planted cycle, models such as GPT-4, Claude 3.5 Sonnet, and Claude 3.7 Sonnet show rapidly diminishing success rates as the cycle length $k$ increases. This shows that although models continue to improve rapidly~\citep{geminiteam2024gemini15unlockingmultimodal,o1-preview,deepseekai2025deepseekr1incentivizingreasoningcapability}, their capacity for exact combinatorial search remains bounded by a ``short-chain'' reasoning frontier.

While this combinatorial reasoning capability can occasionally be improved by fine-tuning a model for a specific global objective~\citep{sanford2024understandingtransformerreasoningcapabilities,merrill2024expressivepowertransformerschain}, such a tuned model would  lose its interpretive flexibility when presented with informal, context-dependent requirements at test time.  This trade-off between interpretive flexibility and global optimality arises broadly in domains like robot navigation~\citep{tellex2011understanding}, molecular design, and automated scheduling~\citep{jobson2024investigating}.

Therefore, problems requiring both interpretive flexibility and enforcement of global constraints exceed the reach of either pure LLMs or pure classical algorithms. For example, in generating a scenic route between two locations, an LLM may expertly evaluate individual route segments for scenic value based on natural language criteria, but it struggles to ensure those segments connect to form a valid, continuous path~\citep{mei2016listen}. Conversely, a combinatorial algorithm easily ensures path connectivity but cannot interpret the qualitative concept of ``scenic-ness.'' Similar complementary trade-offs exist in conference planning~\citep{jobson2024investigating}, where an LLM can semantically analyze abstracts to cluster sessions, but a classical algorithm is required to schedule them into a conflict-free timetable. This interaction is therefore best studied as model-based optimization with expressive black-box generative priors, providing a high-level motivation for the current work.

Within this space, our motivation for MBO comes from inference-time alignment, where a fixed base model is adapted at test time using a reward signal or cost function. Indeed, \topift and \alift are analogous to a test-time training step in LLMs~\citep{TTT}, in that model fitting (that is, fine-tuning) is performed separately for each problem instance. See also~\citep{huang2025inferencealignment,chen2025sets,madaan2023selfrefineiterativerefinementselffeedback} for other empirical inference time alignment procedures.  Our framework provides complementary theoretical bounds (under the coarse learnability assumption) to this line of work.

\section{Empirical Plausibility of \cref{ass:learn}}
\label{sec:empirical}
We now present some empirical evidence to support coarse learnability.
The rigorous theoretical guarantees established in \cref{sec:analysis} and \cref{app:nonconvex_mbo} rely on analytic generative models. Given the difficulty with showing corresponding results for algorithm-LLM interaction, we treat the coarse learnability condition (\cref{ass:learn}) in this section as an empirical heuristic rather than a provable guarantee. The primary contribution of this work is theoretical, and the goal of these experiments is merely to qualitatively study the empirical dynamics of these models. Specifically, we aim to test whether iterative fitting on a very small number of low-cost samples can heuristically shift an LLM's distribution in the spirit of coarse learnability, namely, by demonstrating broader coverage over lower-cost regions.

We use a primitive model, GPT-2, since it will demonstrate that our results are not a byproduct of the complex reasoning capabilities of frontier models, and applies to simple models as well. We also use the heuristic algorithm \topift (\Cref{alg3}),  which does not require estimating the LLM probabilities $\L(s)$ (\cref{ass:llm-prob}) that is required for implementing \alift. This algorithm has the advantage of being simpler and implementable with frontier models where estimating probabilities directly is harder. We therefore present results for this heuristic, noting that the results for \alift are comparable and omitted for brevity. Accordingly, these experiments should be interpreted as qualitative evidence for the learnability intuition underlying \cref{ass:learn} for LLMs rather than as an evaluation of \alift versus \topift. 

\subsection{Line Scheduling}
We focus on a simplified line scheduling problem here, and also present results for the low degree spanning tree problem in \Cref{sec:mst}.  In the line scheduling problem, there are $K$ stations on a line labeled $\{1,\ldots,K\}$. For $1 \le i < K$, the travel time between station $i$ and $i+1$ is $t_i$. Each station $i$ has opening time $o_i$.  The user specifies the intervals $[\ell_i, u_i]$ for $1 \le i \le K$ that capture the lower bound and the upper bound on how long they wish to spend at each station.  The goal is to find a visit duration $v_i$ for each $i \in \{1,\ldots,K\}$ that specifies how much time the user should spend at each station; let $\vec{v}$ be the vector of these durations.  If the user reaches a station before time $o_i$, then they need to wait there until time $o_i$. We constrain all numbers, including those generated by the LLM, to be integers.

We split the problem constraints between a combinatorial algorithm that provides zeroth-order feedback on the objective $d(\cdot)$ and the LLM as follows. 

\begin{description}
\item[Algorithm's Constraints.] 
Given a solution $\vec{v}$, for $i \ge 2$, let $a_i$ be the arrival time at station $i$, and $w_i$ be the waiting time for the station to open; we assume $w_1 = o_1 = 0$. The total wait time is $d(\vec{v}) = \sum_{i=1}^K w_i$, and this is the algorithm's cost (or objective function). 
\item[LLM's Constraints.] The LLM's constraints correspond to the visit time intervals $[\ell_i, u_i]$ for each point $i$. Its cost is the total violation of the visit time computed as $\sum_{i=1}^K \eta_i$, where $\eta_i$ is the distance of the visit time $v_i$ from the interval $[\ell_i, u_i]$. This quantity is $0$ if $v_i \in [\ell_i, u_i]$.
\end{description}

The LLM (GPT-2) is initially fine-tuned to satisfy only the local visit duration intervals, making it ``oblivious'' to the global waiting time constraint. This yields the model $\L_0$. We generate instances with $K = 10$ and where each $t_i, \ell_i, u_i$ is an integer in the range $[1,20]$ with the constraint $\ell_i \le u_i$.  For $i \ge 2$, we set $o_i = o_{i-1} + t_{i-1} + u_{i-1},$ so that there is a solution with wait time $0$ if we set $v_i \ge u_i$ for each $i$. The only solution among these that also has a visit time violation $0$ is the solution that sets $v_i = u_i$ for each $i$. 

In our experiment, we will consider the specific instance where the visit time bounds are set as $\ell_i = 1$ and $u_i = 20$, hence being vacuous, and all travel times are set to $t_i = 10$. For this instance, the optimal visit time vector is $\langle 20, \ldots, 20 \rangle$, and this vector has wait time zero and visit time violation of $0$. If the visit time at a station is $v < 20$, then the solution has to wait $20-v$ steps before the next station opens. 

\begin{figure}[htbp]
\centering
  \begin{minipage}[b]{0.45\textwidth}
    \centering
    \includegraphics[width=\linewidth]{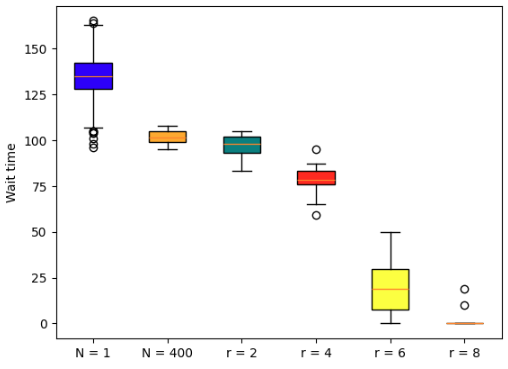}
    \par\medskip
    {\small (a)}\par
  \end{minipage}%
  \qquad
  \begin{minipage}[b]{0.45\textwidth}
    \centering
    \includegraphics[width=\linewidth]{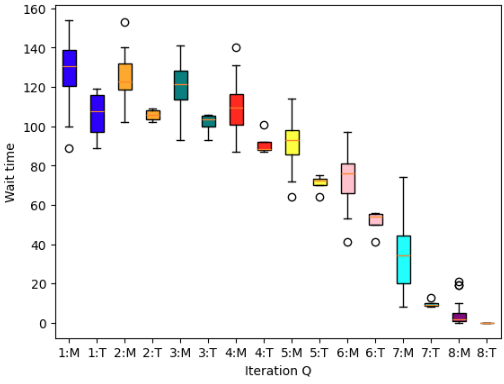}
    \par\medskip
    {\small (b)}\par
  \end{minipage}
\caption{Box-plots of algorithm's cost for the setting where the LLM's visit time constraint is $[1,20]$. In (a), for $20$ runs of \topift,  the left two plots are the distribution of the algorithm's cost for {\sc Best-of-LLM}$(N)$ baseline, while the right four plots are for \topift after $r$ iterations. In (b), for one run of \topift, for different values of iteration $r$, the box-plot ``$r:M$'' is the distribution of the $m \cdot M = 48$ samples generated by the previous model, while ``$r:T$'' is the distribution of $m = 4$ samples among these which have lowest algorithm's cost (waiting time), and which are used for fine-tuning the new model. \label{fig:bothplots} \label{fig:nobox} \label{fig:nobox2} \label{fig:nobox0}
}
\end{figure}

In \Cref{fig:nobox}(a), we show the box plot, for $20$ runs of the algorithm, of the algorithm's cost (total wait time) for the {\sc Best-of-LLM} baseline that draws $N$ samples from $\Lzero$ and outputs the solution with minimum  cost $d(\cdot)$. We plot this for $N = 1$ and $ N = 400$ samples. Note that the median algorithm's cost for {\sc Best-of-LLM} for $N = 400$ is at least $100$. We observe that for any station, if $v$ is the visit time assigned to the station in a sample from $\Lzero$, then assuming this is a uniform distribution on $[1,20]$, we would have $\E[v] = 10.5$, so that the expected wait time is $(20 - \E[v]) \cdot (K-1) = 85.5$. This roughly agrees with the baseline. Further note that $\Pr [v = 20] \le 0.05$. This is because the LLM is initially only aware of the bound $v \in [0,20]$, and hence, is likely to choose an integer visit time uniformly at random from this range. It is unaware of the global waiting time constraint.  Assuming the dimensions behave independently, this means $\Pr[\mbox{Wait time } = 0] \le 10^{-10}$, so that merely sampling from $\Lzero$ is very unlikely to find a solution with optimal waiting time.

We then compare this to \topift (\Cref{alg3}) that fine-tunes $\L$ iteratively using zeroth-order feedback about $d(\cdot)$.  For \topift, we set the parameters as  $m = 4$, $M = 12$, and $Q = 8$, so that the total number of samples used is comparable to $N = 400$ used by the baseline. We consider $20$ independent runs of the algorithm, and  in \Cref{fig:nobox}(a), we show the Box plot of the best waiting time obtained after $r$ iterations of the outer loop of \topift. Though the {\sc Best-of-LLM} baseline yields waiting time (objective $d(\cdot)$) of at least $100$ even with $N = 400$ samples, we note that \topift improves on this even with $r=4$, generating the optimal visit-time vector almost always (wait time = 0) at $r = 8$ iterations. 


In \Cref{fig:nobox2}(b), we show, for one run of the algorithm, the distribution of the $m \cdot M = 48$ samples generated by the model $L_{r-1}$ for $1 \le r \le 8$, and the distribution of the best $m = 4$ samples (according to the algorithm's cost) among these used for fine-tuning $L_r$. Note that initially the fine-tuning samples move the model only slightly, while in later iterations the fine-tuned model produces solutions with substantially lower algorithmic cost than the fine-tuning data itself. Thus, the update is not merely memorizing the elite set, but shifting the model's distribution further toward the low-cost region. This behavior is qualitatively analogous to \cref{ass:learn}: a learner trained on a small sample from a target distribution produces a proposal that covers that target directionally better than the raw sample alone would suggest.\footnote{We note that the same effect cannot be observed with only one round of fine-tuning --- if we set $Q = 1$, $m = 30$, and $M = 12$, and take the best waiting time of the samples from $\L_0$ and $40$ samples from $\L_1$ (so that the total number of samples remains $400$), we observe the waiting time is $81$. This shows the advantage of the iterated framework.} While \Cref{fig:nobox2}(b) does not verify the density-ratio conclusion of \cref{ass:learn}, it provides empirical evidence for the kind of coarse, mass-covering extrapolation that the assumption posits.

This phenomenon, where the model collapses to the optimal solution~\citep{shumailov2024ai} demonstrates a remarkable ability of the LLM to extrapolate beyond the limited fine-tuning data. The iterative guidance from the zeroth-order feedback, even with a small sample size, enables the LLM to learn complex combinatorial constraints that it was initially unaware of. 

\subsection{Low Degree Spanning Tree} 
\label{sec:mst}

In the low degree spanning tree problem, the input is a graph with $n = 16$ vertices numbered $0,1,\ldots,15$. We start by including the Hamiltonian path that includes all edges of the form $(i,i+1)$ for $i \in \{0,1,\ldots,14\}$. Next, for every pair of vertices, we add an edge between this pair with probability $p = 0.4$, removing duplicates. Note that the expected degree of a vertex lies in $[6,8]$, and the graph has approximately $60$ edges in expectation.

The goal is to output a spanning tree minimizing the number of vertices with degree larger than two; in the ideal case, a Hamiltonian path.   We split this problem between a combinatorial algorithm that captures the objective $d(\cdot)$ and LLM as follows:

\begin{figure}[htb]
  \centering

  \begin{minipage}[b]{0.28\textwidth}
    \centering
    \fbox{\includegraphics[width=\linewidth]{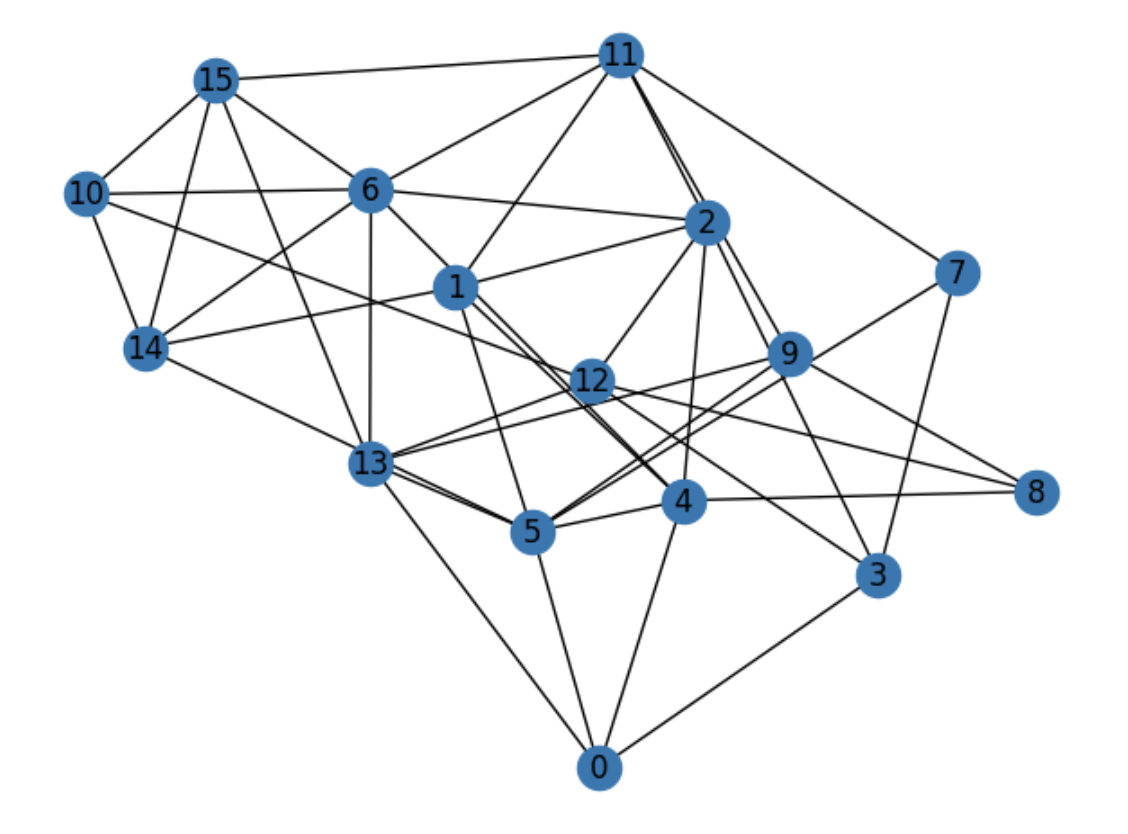}}
    \par\medskip
    {\small (a) Input graph.}
  \end{minipage}
  \qquad
  \begin{minipage}[b]{0.28\textwidth}
    \centering
    \fbox{\includegraphics[width=\linewidth]{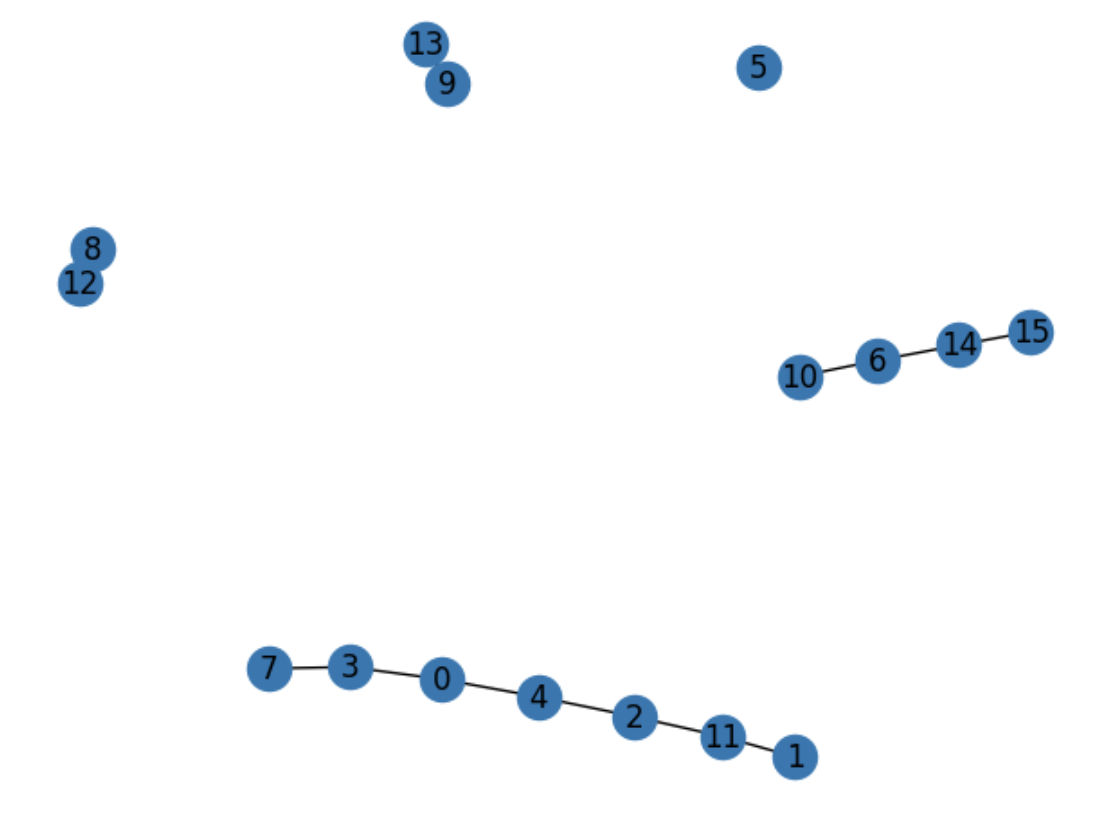}}
    \par\medskip
    {\small (b) {\sc Best-of-LLM}, $N=1$.}
  \end{minipage}
  \qquad
  \begin{minipage}[b]{0.28\textwidth}
    \centering
    \fbox{\includegraphics[width=\linewidth]{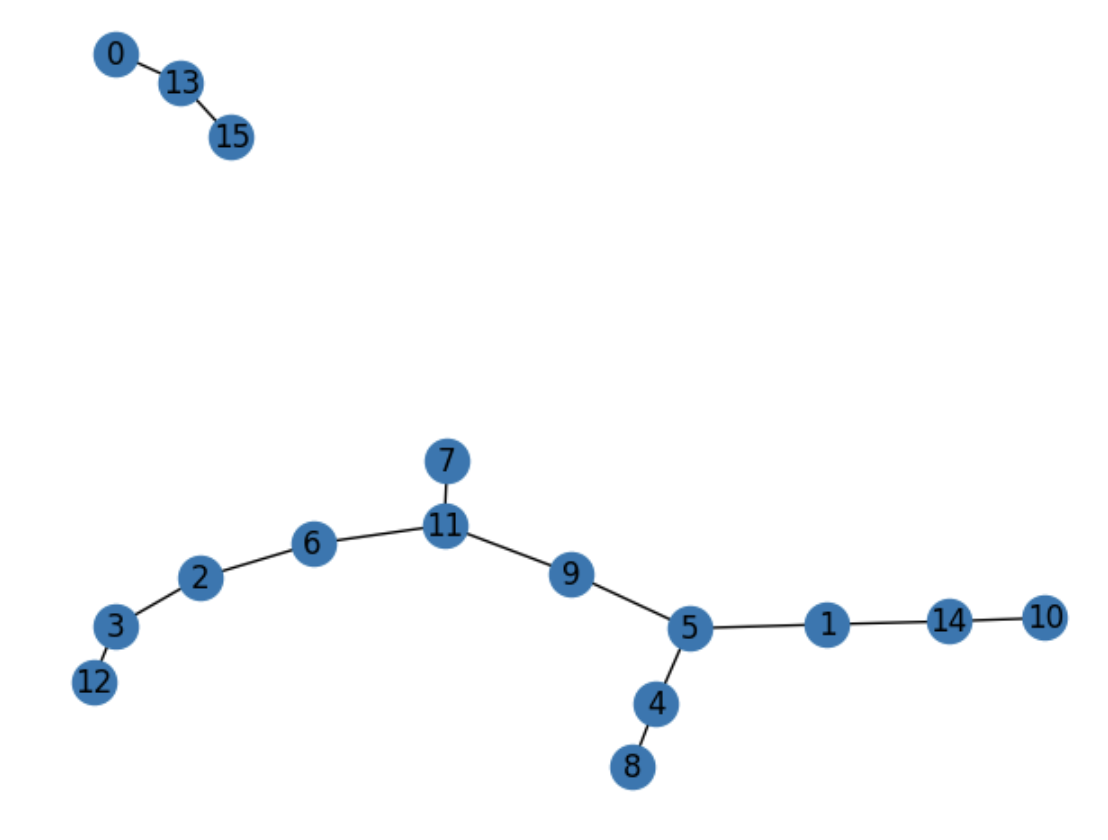}}
    \par\medskip
    {\footnotesize (c) {\sc Best-of-LLM}, $N=600$.}
  \end{minipage}

  \par\medskip

  \begin{minipage}[b]{0.28\textwidth}
    \centering
    \fbox{\includegraphics[width=\linewidth]{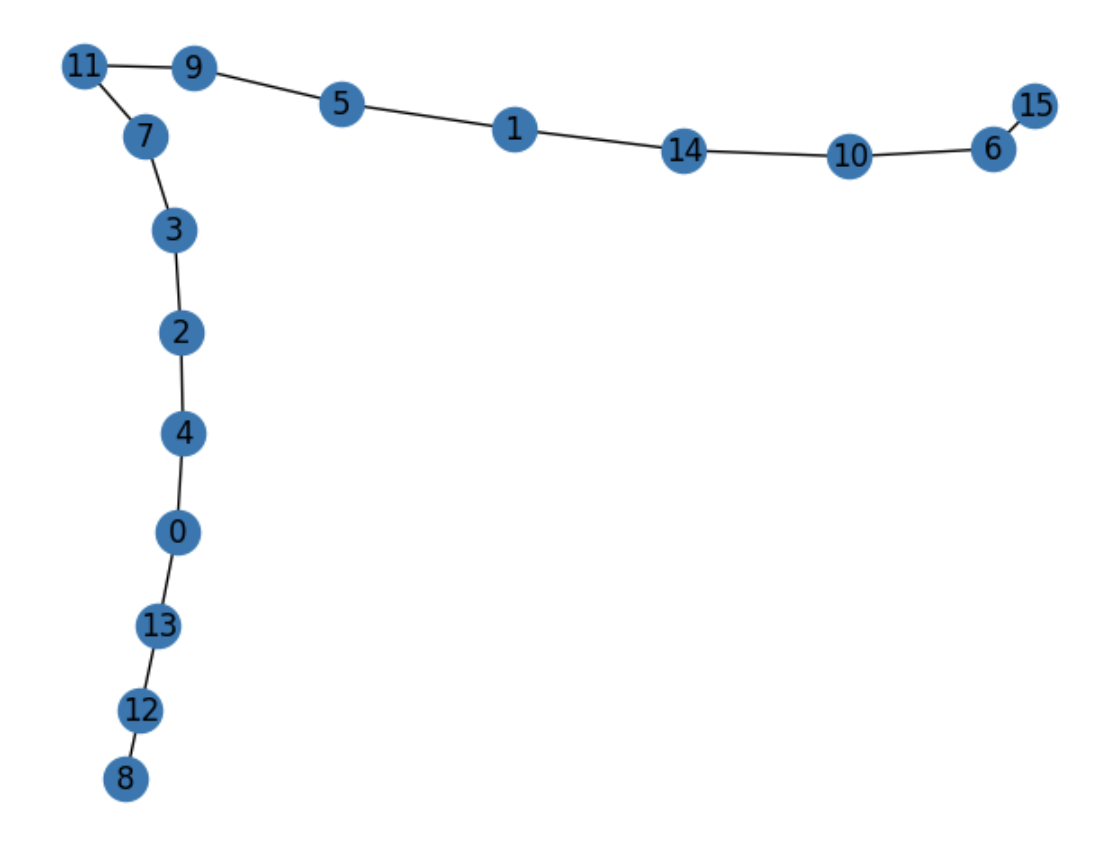}}
    \par\medskip
    {\small (d) \topift\ for $Q=3$.}
  \end{minipage}
  \qquad
  \begin{minipage}[b]{0.28\textwidth}
    \centering
    \fbox{\includegraphics[width=\linewidth]{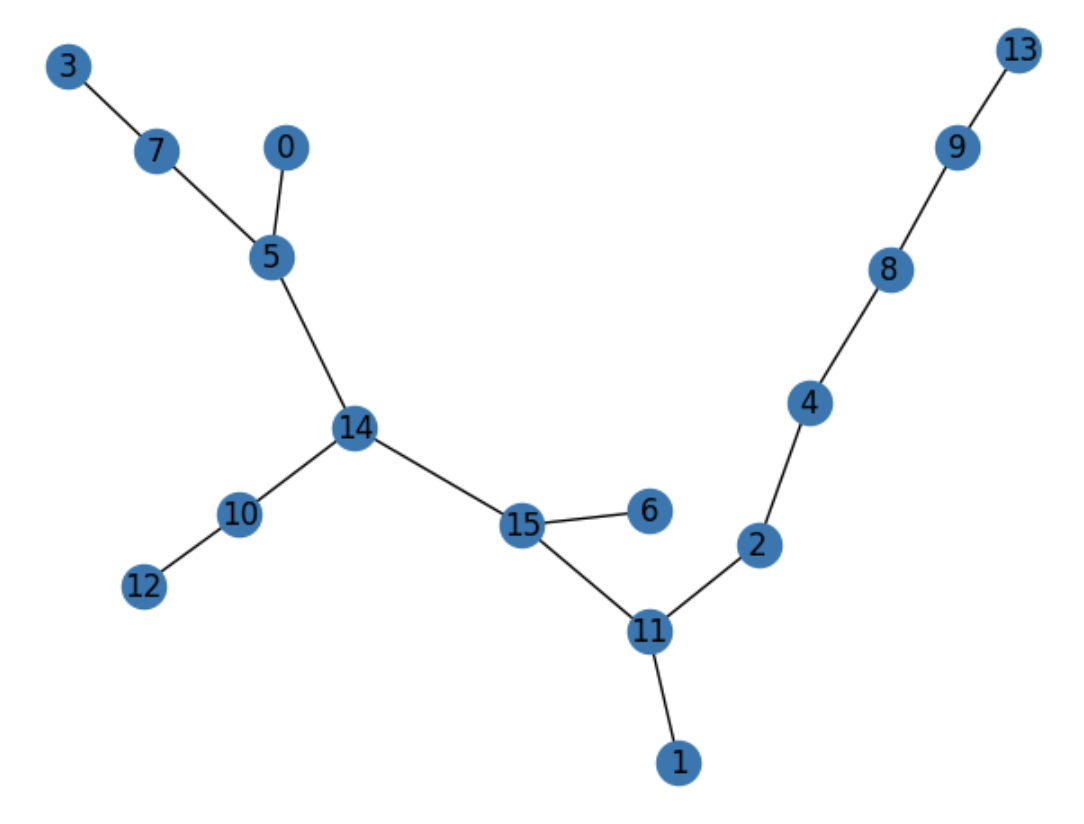}}
    \par\medskip
    {\small (e) {\sc Best-of-ALG}, $N=1$.}
  \end{minipage}

  \caption{ \label{fig:mstfig1}
    \label{fig:mst_graphfig} 
    \label{fig:mst_basefig} 
    \label{fig:mst_bestofLLMpic} 
    \label{fig:mst_FTfig}  
    \label{fig:mst_randomfig} 
    Input graph and outputs of \topift\ and the various baselines.  
    Note that (b) is simply the output of the base model $\L_0$.  
    The output of {\sc Best-of-ALG} for $N = 600$ and random spanning trees is comparable to (e), which means the model $\L_0$ assigns comparable  probabilities to different random spanning trees.
  }
\end{figure}

\begin{description}
\item[Algorithm's Cost.] 
The algorithm simply wants to output a spanning tree. 
Given a solution $x$ as a list of edges, its cost $d(x)$ is the number of connected components induced by $x$ minus $1$, so that the optimal cost is $0$, and the maximum possible cost is $n-1 = 15$. 
\item[LLM's Cost.] These correspond to the local node-wise degree constraints. We measure the cost of the LLM's solution (set of edges) as the number of vertices whose degree is greater than $2$.
\end{description}

We fine-tune GPT-2 on forests with maximum degree two, yielding the base model $\L_0$. We then run the \topift heuristic (\cref{alg3}) with $m = 4$, $M = 50$, and $Q = 3$. 
Note that the total number of samples is $N = m \cdot M \cdot Q = 600$. We compare to two baselines: In the {\sc Best-of-ALG} baseline, we generate $N$ random spanning trees (all with cost $d(\cdot) = 0$) and compute the probability assigned by $\L_0$ to each of them, choosing the best one. In the {\sc Best-of-LLM} baseline, we generate $N$ solutions from  $\L_0$ and score them according to the cost $d(\cdot)$, choosing the best one.



In \cref{fig:mstfig1}, we illustrate the solutions found for a specific random test instance by the different baselines and \topift.  Note that samples from $\L_0$ satisfy the degree constraints, but are disconnected. {\sc Best-of-LLM} for $N = 600$ (\cref{fig:mst_bestofLLMpic}(c)) improves connectivity but still finds a solution that is not only disconnected, but has degree violations. On the other hand, \topift finds a Hamiltonian path\footnote{This does not happen for all $15$ test instances; however, \topift finds one connected component in all instances.}, hence achieving optimum cost for both the algorithm and LLM. Finally, note that random spanning trees (\cref{fig:mst_randomfig}(e)), though connected, have many degree violations. (The latter is also essentially what {\sc Best-of-ALG} for $N = 1$ and $N = 600$ generate.) This visually shows that \topift is incorporating both the algorithm's and the LLM's constraints in a non-trivial fashion, indeed, finding the ``optimum'' solution on this instance. This showcases the power of the LLM in learning a complicated combinatorial constraint from a few samples, which aligns with coarse learnability.

\section{Conclusion}
\label{sec:conclusion_future}


Our framework opens exciting avenues for future research. An ideal goal would be to prove a condition akin to coarse learnability for general LLMs, and a particularly compelling direction is to formally characterize the class of coarse learners and determine under what conditions modern LLMs fall into this category. For instance, which architectural features, training regimes, or fine-tuning strategies enable these models to satisfy the coarse learnability assumption? Addressing this question may require synthesizing insights from recent work on transformers and chain-of-thought reasoning through the lens of circuit complexity~\citep{merrill2024expressivepowertransformerschain}.   
Another direction is to explore more powerful optimization oracles, such as those providing gradients of the log-probability, which could enable the use of Langevin MCMC in place of Metropolis--Hastings, potentially enhancing the efficiency and scalability of the framework. Finally, it would be interesting to extend \cref{app:nonconvex_mbo} to work with weaker assumptions than quadratic bounding, and to extend \alift to remove \cref{ass:llm-prob}. Collectively, these directions suggest a rich landscape for integrating statistical learning theory with combinatorial optimization, pointing toward a principled foundation for adaptive generative models.


\appendix
\crefalias{section}{appendix}
\crefalias{subsection}{appendix}
\section{Additional Discussion on \alift and Coarse Learnability}
\label{sec:discussion}

\subsection{Connecting \topift to \alift}
\label{app:topift}
Recall \topift from \cref{alg3}. We now describe a smooth approximation of \topift that avoids the hard selection of the $m$ smallest-cost samples, and show that this naturally leads to a procedure akin to \alift.

Towards this end, we position \topift within a framework in which the algorithm tries to make the distribution $\L_r$  a progressively good approximation to the  distribution $p_T^*$.  Unfortunately, obtaining a closed form for $\L_r$ in \topift seems difficult because we choose the samples with the smallest $d(\cdot)$, i.e., hard-min.  To circumvent this difficulty, we will use a soft-min distribution to approximate this step; this leads to a simulated annealing algorithm that we describe next. 

Note that the construction of $S'_{r}$ in \cref{alg3} can be split into two steps: 
 \begin{enumerate}[nosep]
     \item We generate $s$ from the distribution $\L_{r-1}$. If this is repeated $m \cdot M$ times, this corresponds to generating $S_r$. 
     \item We select the samples from $S_r$ with a minimum $d(\cdot)$. This can be smoothly approximated by keeping each $s \in S_r$ with probability $e^{-\tau_r \cdot d(s)}$. This is a ``soft-min'' that favors solutions $s$ with smaller $d(s)$. The parameter $\tau_r$ can be chosen so that if $D = \max_{s \in S} d(s)$, then $\tau_r = \frac{\ln M}{D}$. This ensures that any $s \in S$ is sub-selected in this step with probability at least $\frac{1}{M}$, so that $\E[|S'_r|] \ge m$. 
 \end{enumerate} 
 
Therefore, the sampling distribution of $S'_r$ is proportional to $\L_{r-1}(s) \cdot e^{-\tau_r \cdot d(s)}$.  If we assume that $\L_r$ is faithfully learned from these samples, we obtain $\L_r(s) \propto \L(s) \cdot e^{-\sum_{t = 1}^r \tau_t \cdot d(s)}$.  We can now choose the number of iterations $Q$ so that $\sum_{t = 1}^Q \tau_t \approx T$, which leads to the distribution $p_T$. This yields a smoothed variant of \topift, analogous to simulated annealing, where the temperature $\tau = \sum_{t=1}^r \tau_t$ is gradually increased across iterations and a model $p_{\tau}$ is learned. The gradual increase ensures that the set $S'_r$ at each iteration remains sufficiently large.

A key challenge is that in each iteration, the model $\L_r$ only approximates the sampling distribution $S'_r$. This approximation can be quite crude, since the distribution $p_{\tau}(s) \propto \L(s) \cdot e^{-\tau \cdot d(s)}$ encodes combinatorial constraints that the model cannot fully capture, even with many samples. Consequently, errors can accumulate multiplicatively across fine-tuning iterations, making it difficult to formally prove convergence for the simple heuristic. We address this issue by incorporating a Metropolis--Hastings step to select the top samples. This allows the approximately learned model $\L_{r-1}$ to serve as an efficient proposal distribution, while the acceptance rule ensures sampling from the true target distribution $p_{\tau}$, preventing error accumulation. This leads to the \alift algorithm.

\subsection{\alift vis--a--vis Model Reference Adaptive Search (MRAS)}
\label{sec:mras_connection}
Our analysis of \alift can be interpreted as a finite-sample robustification of the model reference adaptive search (MRAS) framework \citep{hu2007model} via the Metropolis--Hastings proposal step. In the MRAS framework, one defines a sequence of ideal target distributions, $g_k(s)$, and updates a parametric model $f(\cdot; \theta_k)$ to approximate $g_k$ by minimizing the KL-divergence $D_{\text{KL}}(g_k \| f_\theta)$. In our instantiation, $g_k(s) \propto \model_0(s) e^{-\tau_k d(s)}$.

Standard MRAS theory guarantees $f(\cdot; \theta_k)$ converges to a point mass at the globally optimal solution for $d$. However, this analysis relies both on exactly updating the parameter $\theta$ via infinite samples as well as the specific structural properties of natural exponential families to ensure the optimum always remains within the support of this generated distributions \citep{hu2007model}. Therefore, this theory does not provide explicit sample complexity bounds. In the finite-sample regime, projecting a complex target $p_\tau$ onto a generative model introduces empirical errors. Without specific statistical guarantees, there is no assurance that a polynomial number of samples suffices to maintain coverage of the optimal region, and as mentioned before, this aspect may break the algorithm.

Our framework identifies the statistical assumption (coarse learnability) and algorithmic modification (utilizing the coarse generative model as a proposal for Metropolis--Hastings rather than sampling from it directly) needed for achieving polynomial sample complexity in classical MBO. While methods like MRAS rely on the rigid geometry of exponential families to guarantee support preservation, \cref{ass:learn} requires that the learner captures the target's mass within a polynomial factor. Our framework therefore shifts the burden from architectural correctness (choosing the right parametric family) to statistical expressivity (learning a cover), making the theory applicable to settings where exact parametric alignment is impossible. We presented a simple case of such model mis-match in \cref{app:miss}.

\section{Prompts for Empirical Results}
\label{app:prompts}
\begin{promptbox}{Finding Cycle in a Graph}
"""Identify a cycle of length exactly \{k\} in this undirected graph.
Edges: \{edges\}

Think step by step. Finally the last line of your response must be the final output.
The last line MUST be a list like [0,1,2,...,0] or 'No cycle found'. No other text:"""
\end{promptbox}

\begin{promptbox}{Line Scheduling}
"""
    There are \{n\} museums on a line. They are numbered from 0 to \{m\}.
    I'm currently located at museum 0 and the current timestamp is 0.
    I want to visit all these museums one by one in a sequence. Each museum has an opening time.
    If I reach a particular museum before it opens then I may have to wait. The opening times for the museums are as follows: 
    
    [\textbf{opening times go here}]  
    
  In addition, the following list of \{m\} numbers contains the time to travel from museum i to i+1.
  So the first number is the time to travel from 0 to 1 and so on: 
  
  [\textbf{travel times go here}]
  
  
  
  Finally, I have certain constraints in terms of the minimum and maximum amount of time I want to visit each museum.
  This is described as the following list of arrays: 
  
  [\textbf{constraints go here}]
  
  Give me a schedule in terms of a list of \{n\} numbers
  describing how much time I should spend at each place so that all my constraints are satisfied and at the same time
  my total wait time is as little as possible. Do not use code. Simply output the list of \{n\} numbers (one per line) and nothing else."""
\end{promptbox}

\begin{promptbox}{Spanning Tree}
"""You are given a graph with vertices labeled from 0 to 
\{num\_vertices-1\}. Each line below lists an edge of the graph as (i,j).

 [\textbf{edges go here}]

  Your goal is to output a list of H of a subset of the edges that form a spanning tree, i.e., the subgraph induced by H should be connected. Furthermore, each vertex should
  appear in at most 
  \{deg\} times in the list. Simply output the list of edges and nothing else. Format your answer by producing one edge per new line."""
\end{promptbox}

\section{Acknowledgment of Generative AI Use}
\label{app:ai_use}
We utilized Gemini 3.1 Pro to assist with (1) identifying connections to model-based optimization in \cref{sec:mras_connection}; (2) identifying relevant literature to simplify the proof of \cref{thm:main_iter}; (3) improving comparisons to prior work by finding citations; and (4) generating the code for the experiments in \cref{sec:empirical_10d}. We also benefited from discussions with the model, which helped suggest possible approaches to proving the results in \cref{app:nonconvex_mbo,app:kde}. All results, including citations and algebra, were verified by the authors, who take full responsibility for the paper’s accuracy and contributions.

\end{document}